\pdfoutput=1
\documentclass[preprint,12pt]{elsarticle}

\usepackage{amsmath,amssymb,amsthm}
\usepackage{tikz}
\usepackage{float}
\usepackage{needspace}
\usepackage{array}
\usepackage{booktabs}
\usepackage{url}
\usepackage{mathpartir}
\usepackage[cal=boondoxo]{mathalfa}   %
\usepackage[hidelinks]{hyperref}

\theoremstyle{plain}
\newtheorem{theorem}{Theorem}
\newtheorem{lemma}{Lemma}
\newtheorem{proposition}{Proposition}
\newtheorem{assumption}{Assumption}
\theoremstyle{definition}
\newtheorem{definition}{Definition}

\newcommand{\Obs}{\mathit{obs}}
\newcommand{\safe}{\mathit{safe}}
\newcommand{\Post}{\mathit{Post}}
\newcommand{\may}{\mathcal{m}}
\newcommand{\must}{\mathcal{M}}
\newcommand{\kn}{\kappa}

\journal{Artificial Intelligence}

\begin{document}

\begin{frontmatter}

\title{Learning Inside Types: Epistemic Typestates for Safe Multi-Agent Reinforcement Learning}

\author{Lorenzo Bacchiani\corref{cor1}}
\ead{lorenzo.bacchiani2@unibo.it}
\author{Mario Bravetti}
\author{Giovanni Pau}
\author{Roberto Girau}
\cortext[cor1]{Corresponding author}
\affiliation{organization={University of Bologna}, city={Bologna}, country={Italy}}

\begin{abstract}
Type disciplines make programs safe by construction: components checked against each other's types compose into
programs that cannot go wrong in the ways the types exclude. Types suit reinforcement learning: a type constrains
what an agent may choose, not how it chooses, so the guarantee holds for every policy a learner produces. We introduce
epistemic typestates, a type discipline for safe multi-agent reinforcement learning. An agent's type is a protocol
whose states record its beliefs, the worlds it cannot rule out from its own observations and actions, and permits
only actions that are safe in every world of the belief, whatever the others do within their public types. Agents
learn inside their types, which are built and checked one agent at a time and run without communication. We prove
type soundness and team safety, give a synthesis that is sound and complete, build types from
guards when worlds are infinite, and show that shields are a special case. All proofs are mechanised in Rocq. On nine
benchmarks, no typed learner causes a safety violation, and task performance does not differ significantly from that
of two recent multi-agent shielding methods. Where a contract-based method fails at five and
ten agents, types for ten agents are built in under a second. In the two continuous benchmarks, Lagrangian learners
cause thousands of violations in every training run.
\end{abstract}

\begin{keyword}
Safe reinforcement learning \sep Multi-agent reinforcement learning \sep Typestates \sep Behavioural types \sep
Shielding \sep Partial observability
\end{keyword}

\end{frontmatter}

\section{Introduction}\label{sec:intro}

Type disciplines make programs safe by construction. A type states what a component may do, a checker verifies each
component against the types of the others, and in a sound discipline the components that pass the check compose into a
program that cannot go wrong in the ways the types exclude. Typestates~\cite{StromY86} add protocols, so that the
operations a component may perform depend on its state, and typestate checkers verify such protocols
statically~\cite{BacchianiBGMR22}. Reinforcement learning (RL) agents that act in a shared space need this kind of
guarantee: every action that a learner tries, in training as in execution, must be safe. A type discipline offers it
in a form suited to learning. A type constrains what an agent may choose, not how it chooses, so the guarantee holds
for every policy that the learner produces. Agents rely on each other's types, which act as
interfaces~\cite{AlfaroH01}, rather than on each other's policies, which keep changing during training. Each agent is
checked separately, before learning starts, so the checks remain per agent as teams grow. 
With several agents, whether an action is safe also depends on what the other agents do and on parts of the world that
the acting agent cannot sense: an agent's safety depends on another that it
cannot see. Shields~\cite{AlshiekhBEKNT18} are an established way to keep learners safe. With several agents, they
handle such dependencies through a model of the whole team: they are synthesised over joint states or their
projections~\cite{ElSayedAlyBAETF21,BrorholtLS25}, decomposed from a centralised shield~\cite{MelcerAT24}, projected
from a global machine~\cite{YangY26}, or certified together over a global model~\cite{AdalatHB26}. These
constructions require a finite joint model or abstraction, and whether an agent can then run its shield from its own
information, without communication, is a property of the derived shield. %

We start instead from what each agent knows. An \emph{epistemic typestate} is a protocol for one agent whose states record
its beliefs: the worlds that the agent cannot rule out from its own observations and actions. Like a classic typestate~\cite{StromY86,BacchianiBGMR22,BacchianiBGMR24}, it states which actions are permitted in each state, and permits only actions that
are safe in every world of the belief and leave a continuation in the protocol. The agent reasons about the other
agents only through their public types, which state what they may do and the safety requirements they guarantee, while
declarations let an agent commit to actions that others can rely on, as obligations to take an action or restrictions
to a set of actions.
Each type is built and checked for its own agent against the public types of the others. The learner itself is not
checked but confined to the actions its type permits, so every policy it learns conforms by construction. Each agent
runs its type on its own observations, so execution needs no communication.
We implement epistemic typestates in an engine that builds, analyses and runs types. Its input is the model of each
agent, e.g., its worlds, their dynamics, its observations, together with the public types
of the other agents. The engine synthesises a type when the agent's reachable beliefs are finite,
or builds a type from a guard, i.e., a rule that gives the permitted actions from the agent's observation. This
construction also applies when the worlds are infinite, as with continuous positions. Its
static analyses report, for instance, which observations a type needs and what it loses against an exact-knowledge
reference, and the finite types it builds are certified by an independent checker proved sound in Rocq\footnote{https://rocq-prover.org}. Finally, the
engine runs each type to supply its learner's action mask.

We evaluate learning inside epistemic typestates on nine benchmarks: six in the configurations of Adalat
et al.~\cite{AdalatHB26}, with Car Platoon also in the version of Brorholt et al.~\cite{BrorholtLS25}, Traffic
Junction~\cite{SinghJS19}, and two with continuous worlds, MPE~\cite{LoweWTHAM17} and VMAS~\cite{BettiniKBP24}.
Learners trained inside the types are compared with those protected by the shields of Adalat et al.\ and
Brorholt et al., with unshielded learners in Traffic Junction, and with Lagrangian learners~\cite{GuEtAl23} in the
continuous benchmarks. We measure safety in training and evaluation, task performance, construction time and the
freedom left to the learner.

Our contributions are the following.
\begin{itemize}
\item Epistemic typestates, an expressive and scalable type discipline that makes multi-agent RL systems safe by
construction (Section~\ref{sec:ets}). We prove type soundness and team safety, show that shields are a special case,
build types from guards when the worlds are infinite, and extend knowledge-based synthesis for games with imperfect
information~\cite{Reif84,ChatterjeeDHR07} to declarations, keeping it sound, complete and, outside obligations,
maximally permissive.
\item The engine, and a mechanisation in Rocq of all our formal results, including the soundness of the
independent checker (Sections~\ref{sec:ets} and~\ref{sec:evaluation}).
\item An evaluation on nine benchmarks (Section~\ref{sec:evaluation}). No typed
learner causes a safety violation, and task performance does not differ significantly from that of two recent
multi-agent shielding methods in all comparisons but one, although one of them reads state that no agent can sense in
three benchmarks. Where Adalat et al.'s construction fails at five and ten agents, types
for ten agents are built in under a second. In the continuous benchmarks, Lagrangian learners cause thousands of
violations in every training run.
\end{itemize}

The remainder of the paper is organised as follows. Section~\ref{sec:running} presents the running example, Section~\ref{sec:background} the background and
Section~\ref{sec:related} related work.
Section~\ref{sec:ets} develops epistemic typestates, Section~\ref{sec:evaluation} evaluates them,
Section~\ref{sec:discussion} discusses the results and the limitations, and Section~\ref{sec:conclusion} concludes.

\section{Running example}\label{sec:running}

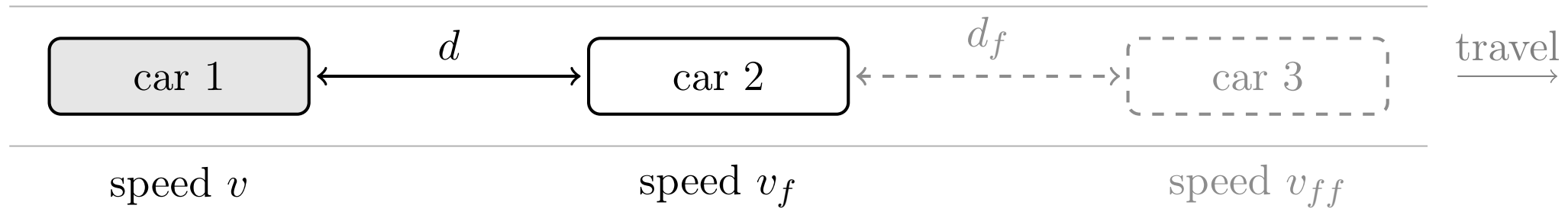
\begin{figure}[htbp]
\centering
\resizebox{\linewidth}{!}{%
\begin{tikzpicture}[scale=0.9, every node/.style={font=\small}]
  \draw[black!30] (-0.4,0.7) -- (13.8,0.7);
  \draw[black!30] (-0.4,-0.7) -- (13.8,-0.7);
  \draw[thick, rounded corners=3pt, fill=black!10] (0,-0.38) rectangle (2.6,0.38);
  \node at (1.3,0) {car~1};
  \draw[thick, rounded corners=3pt] (5.4,-0.38) rectangle (8.0,0.38);
  \node at (6.7,0) {car~2};
  \draw[thick, dashed, black!45, rounded corners=3pt] (10.8,-0.38) rectangle (13.4,0.38);
  \node[black!45] at (12.1,0) {car~3};
  \draw[<->, thick] (2.68,0) -- node[above] {$d$} (5.32,0);
  \draw[<->, thick, dashed, black!45] (8.08,0) -- node[above, black!45] {$d_f$} (10.72,0);
  \node at (1.3,-1.1) {speed $v$};
  \node at (6.7,-1.1) {speed $v_f$};
  \node[black!45] at (12.1,-1.1) {speed $v_{ff}$};
  \draw[->, black!50] (14.1,0) -- (15.1,0) node[midway, above] {travel};
\end{tikzpicture}
}
\caption{Car~1 senses its speed $v$, the speed $v_f$ of car~2 and their gap $d$ (black). The speed
$v_{ff}$ of car~3 and the gap $d_f$ of car~2 are hidden from it (grey, dashed).}
\label{fig:platoon}
\end{figure}

In the Car Platoon benchmark, cars drive in a single lane behind a lead car that \emph{brakes}, \emph{coasts} or
\emph{accelerates} at random. Every other car is a \emph{follower}, controlled by its own agent. At each time step all
cars act simultaneously: a follower brakes, coasts or accelerates, which changes its speed by $-2$, $0$ or $+2$ within
$[0,2]$.
The gap between successive cars changes according to the mean of their speed differences before and after the step.
Each follower must always keep its gap to the preceding car strictly between $0$ and $20$: a gap of $0$ or less is a
collision, and a gap of $20$ or more means that the follower fell behind. The platoon starts at rest, with all gaps
$10$. No communication is allowed. Figure~\ref{fig:platoon} shows car~1 following car~2, which follows car~3.
For car~1, $5$ quantities matter for safety: its own speed $v$, the speed $v_f$ of car~2, its own gap $d$, the speed
$v_{ff}$ of car~3, and the gap $d_f$ of car~2. Notice that $v_{ff}$ and $d_f$ matter because they constrain what
car~2 may do. For instance, suppose car~2 drives at speed $2$ and car~3 is at rest, only $3$
units in front of it. Then car~2 must brake: if it does not and car~3 stays at rest, the gap between
them drops to $1$ and car~2 can no longer avoid a collision. So car~2's behaviour, and hence car~1's
safety, depends on hidden quantities.

A \emph{configuration} gives a value
to each quantity: writing $(v,v_f,d,v_{ff},d_f) = (0,2,11,2,10)$, it says that car~1 is at rest,
car~2 drives at speed $2$ eleven units ahead of it, and car~3 drives at speed $2$ ten units further.
Car~1's sensors read its own speed, the speed of car~2 and its own gap, that is $v$, $v_f$ and $d$. Car~3
is out of sight, so $v_{ff}$ and $d_f$ are hidden. Configurations that agree on $v$, $v_f$ and $d$ look the
same to car~1: $(0,2,11,2,10)$ and $(0,2,11,0,9)$, where car~3 is at rest nine units ahead of car~2,
both read $(0,2,11)$.

\section{Background}\label{sec:background}

\subsection{Multi-agent reinforcement learning}\label{sec:bg-marl}

In multi-agent reinforcement learning (MARL), several agents learn by trial and error to act in a shared
environment. When each agent senses only part of the environment, the team is commonly modelled as a decentralised
partially observable Markov decision process (Dec-POMDP)~\cite{OliehoekA16}. At each time step, every agent $i$
receives an observation $o_i$ of the current state $s$ of the environment and chooses an action $a_i$ from its set of
actions $A_i$. The joint action $(a_1,\dots,a_n)$ then moves the environment to a next state, drawn from a probability
distribution, and every agent receives a reward. A policy $\pi_i$ maps the history of the agent's observations to a
probability distribution over $A_i$, and the agents seek policies that maximise the expected discounted sum of the
rewards, the return. During training, the agents explore actions and update their policies from the rewards they
receive. During execution, they follow the learned policies. Execution is decentralised when every agent acts on its
own observations, without communicating. Independent learners, such as independent Q-learning (IQL)~\cite{Tan93}
and independent proximal policy optimisation (IPPO)~\cite{DeWittEtAl20}, train each agent as if it were alone,
treating the other agents as part of the environment.

\subsection{Shielding}\label{sec:bg-shielding}

Shields were introduced to enforce safety specifications at run time on reactive systems~\cite{BloemKKW15,
KonighoferABHKTW17} and then on learning agents~\cite{AlshiekhBEKNT18}. A shield is computed before learning from
the safety specification and from a model of the environment, typically an abstraction, that records which states
may follow each state and action, but not their probabilities. Computing it amounts to solving a safety
game~\cite{BernetJW02} between the agent, which chooses the actions, and the environment, which chooses among the
possible next states. The winning region is the set of states from which the agent can keep the system safe forever,
whatever the environment does, and the shield permits, in each winning state, only actions whose possible next
states are all winning. A maximally permissive shield permits all of them: it restricts the agent as little as
possible and forbids an action only if the action could endanger safety. A shield can be placed in two
ways~\cite{AlshiekhBEKNT18}. A pre-emptive shield gives the agent, at each step, the set of safe actions to choose
from. A post-posed shield monitors the action that the agent chooses and replaces it only if it is unsafe. Either
way, no unsafe action is executed, during training as during execution and whatever the agent's policy, as long as
the environment behaves as the model says. Probabilistic shields bound the probability of a violation
instead~\cite{JansenKJSB20}. With several agents, a centralised shield monitors the joint action of all
agents~\cite{ElSayedAlyBAETF21}, whereas per-agent shields each restrict the actions of one agent. 

\subsection{Typestates}\label{sec:bg-typestates}

Type systems ensure that programs do not perform invalid operations, such as calling a method that an object does
not define, but not that operations are performed in a valid order: a file, for instance, must be opened before it
is read or written, and cannot be used after it is closed. The operations available on such objects depend on their
internal state. Typestates~\cite{StromY86,DeLineF04}, which are behavioural
types~\cite{HuttelEtAl16}, make this dependence explicit. A typestate is a protocol that
describes, as a state machine, which operations an object permits in each of its states and the state that each
operation leads to. The next state may depend on the result of the operation, for example on whether a read reached
the end of the file. A typestate checker verifies statically that a program uses every object according to its
protocol, as Mungo~\cite{KouzapasDPG16} and JaTyC~\cite{BacchianiBGMR22,BacchianiBGMR24} do for Java. The check is sound: a program
that passes it never performs an operation that the current state of the object does not permit, so protocol
violations are ruled out before the program runs. 
\section{Related work}\label{sec:related}

Beyond the shielding and typestates of Section~\ref{sec:background}, epistemic typestates draw on safe RL
and knowledge-based control. We review these connections, focusing on the information agents need to act safely
and on how local guarantees compose. We close with the multi-agent shielding methods closest to ours and
compare them with epistemic typestates.

\paragraph{Safe RL}
Safe RL methods either change the optimisation criterion, for instance to bound an expected cost, or change the
exploration process using external knowledge~\cite{GarciaF15}. Brunke et al.~\cite{BrunkeGHYZPS22} survey safe
learning in robotics, from learning-based control to safe RL. Provably safe RL guarantees safety at every step of
training and deployment. Krasowski et al.~\cite{KrasowskiTMSWA23} classify its methods by what happens to an
unsafe action of the learner: it is replaced, projected onto a safe action, or masked, so that the learner cannot
choose it. Learning inside a type is action masking: at each step, the actions permitted by the agent's typestate
form its action set. Policy-gradient updates remain valid when computed for the masked policy~\cite{HuangO22}.
Fulton and Platzer~\cite{FultonP18} combine verified hybrid-system models with runtime monitoring to constrain
learning while the observed behaviour conforms to the model.

\paragraph{Shielding under partial observation}
Single-agent shielding under partial observation already reasons about sets of possible states. Junges
et al.~\cite{JungesJS21} synthesise winning regions over belief supports for POMDPs, and Carr
et al.~\cite{CarrJJT23} integrate such shields with deep RL. Their support estimator updates possible states
from action and observation histories without needing transition probabilities. Their avoidance guarantee tolerates
conservative transition models. Their almost-sure reachability result assumes exact transition support and
fairness of the permitted choices. Mazzi et al.~\cite{MazziCF23} shield online POMCP planning using rules fitted
from belief--action traces and expert templates, including bounds on modelled risk. Our contribution concerns how these beliefs form
protocols with declarations, how protocols are synthesised, and how their safety guarantees compose through public
types.

\paragraph{Safety filters for continuous multi-robot control}
Control barrier functions certify invariance of a safe region. Safety filters can use them to minimally modify a
nominal control input. Wang et al.~\cite{WangAE17} derive decentralised safety barrier certificates for collision
avoidance among robots modelled as double integrators: each robot corrects its own input using the relative
positions and velocities of nearby robots. Qin et al.~\cite{QinZCCF21} learn decentralised neural barrier
certificates jointly with multi-agent policies. Reciprocal collision avoidance~\cite{BergLM08,BergGLM11} lets each
robot take half of the responsibility for avoiding every neighbour. ORCA~\cite{BergGLM11} turns this into
half-planes of admissible velocities computed from the neighbours' positions and velocities. These approaches use
nearby robots' states. Our guard construction
(Proposition~\ref{prop:guard}) is parametric in the dynamics and safety requirement: an observation-based guard
must offer an action throughout a safe region containing the possible starts, and every offered action must keep
all possible successors there. This permits safety to depend on unsensed quantities, such as other agents'
velocities. The formalism allows continuous worlds but retains a finite action set.

\paragraph{Constrained multi-agent RL}
Constrained RL expresses safety through cost budgets. Constrained policy optimisation~\cite{AchiamHTA17}
controls expected cumulative costs during policy improvement. Gu et al.~\cite{GuEtAl23} formulate constrained
Markov games and develop multi-agent constrained policy optimisation and MAPPO-Lagrangian. Their theoretical
policy-update procedure guarantees constraint satisfaction and monotonic reward improvement under its assumptions.
An expected-cost budget does not in general exclude every unsafe transition. Our safety
guarantee applies throughout training and execution to every policy that selects permitted actions, independently
of its reward optimisation.

\paragraph{Cooperation without communication}
For Dec-POMDPs (Section~\ref{sec:bg-marl}), optimal decentralised planning is NEXP-hard already for finite
horizons~\cite{BernsteinGIZ02}. Epistemic typestates leave optimisation to the learner while constraining its
choices through safety requirements and declarations. Much work in multi-agent RL lets agents learn to
communicate~\cite{FoersterADW16,SukhbaatarSF16,SinghJS19}.
We assume no communication channel at all. Social laws~\cite{ShohamT95,HoekRW07} coordinate agents without runtime
communication: restrictions on actions, designed offline, such that agents that obey them avoid undesirable
states. Declarations play a similar role for learning agents, with commitments local to each agent: they are part
of its own typestate, and other agents rely on them through its public type. Commitments between agents are also
central to agent communication~\cite{Singh99}, where commitment-based protocols create and discharge them through
messages~\cite{YolumS02}. Declarations need no messages.

\paragraph{Knowledge and supervisory control}
Reasoning about knowledge in multi-agent systems rests on possible worlds~\cite{FaginHMV95}. Our beliefs are sets
of possible worlds in this sense. Knowledge-based programs~\cite{FaginHMV97} make actions depend on what an agent
knows. Engesser et al.~\cite{EngesserMLSZ25} integrate epistemic logic with RL and use epistemic action conditions
to filter choices through model checking. Alpturer et al.~\cite{AlpturerHM24} derive knowledge-based vehicle
protocols for intersections, establishing safety, liveness and conditional optimality. These works provide direct
precedents for epistemic action selection and traffic safety. Supervisory control under partial
observation~\cite{LinW88} enables or disables events using observed histories. Decentralised supervisory
control~\cite{RudieW92} studies how local controllers jointly enforce global behavioural constraints. Ricker and
Rudie~\cite{RickerR00} explicitly formalise decentralised supervisory decisions through possible-worlds knowledge
and knowledge-based protocols. We extend this connection between information and permitted behaviour to learning
agents, using declarations and public types to state their commitments and compose their guarantees. Our results
guarantee safety and continued availability of an action: they impose no liveness requirement on the learner's choices.

\paragraph{Shielded multi-agent RL}
For several agents, individually permitted choices must also be safe when combined. Bharadwaj
et al.~\cite{BharadwajBDKT19} synthesise shields that correct the agents so that a global specification always
holds, minimising the cost of the corrections. ElSayed-Aly et al.~\cite{ElSayedAlyBAETF21} develop centralised shields
over joint actions and factored shields with runtime coordination between regions. Xiao et al.~\cite{XiaoLD23}
let shields split and merge as the agents' states change, with a world model learned during early exploration and
high-probability guarantees. Melcer et al.~\cite{MelcerAT22} decompose a centralised shield into per-agent shields
that need no communication, for agents that all compute the same abstract label of the state. They later extend
the decomposition to agents with different partial observations~\cite{MelcerAT24}. This construction uses a
finite joint abstraction and also supports bounded observation histories, and the underlying environment can be
infinite. Yang and Yoshida~\cite{YangY26} compile a shield process algebra through a global Mealy machine into
local machines whose states represent beliefs about the global machine, and compute best- and worst-case safety
probabilities over all policies by probabilistic model checking. These approaches establish that both
history-dependent local shielding and communication-free execution are possible.

Brorholt et al.~\cite{BrorholtLS25} construct one shield per agent over projections of a finite state space,
using ordered assume-guarantee reasoning: each agent's guarantee may rely only on the guarantees of the agents
before it, which avoids circular reasoning. Their shields are memoryless, and execution requires no
communication. However, continuous systems are future work.

Adalat et al.~\cite{AdalatHB26} use local temporal contracts for a shared global safety objective and certify
them together over a finite global model, as profiles that assign a contract to each agent. Their simultaneous
certification supports circular assumptions.
Decentralised execution requires that each local mask be determined by the active contract and the agent's local
observation history. Contract monitors can supply memory. Selecting contracts during training involves
synchronisation at episode or dwell boundaries, while execution of a fixed certified contract needs no
communication when this information condition holds.

\paragraph{Positioning}
Table~\ref{tab:positioning} compares the runtime information, construction models and composition assumptions of
these multi-agent shielding methods with epistemic typestates.
\begin{table}[!t]
\centering
\footnotesize
\setlength{\tabcolsep}{3pt}
\renewcommand{\arraystretch}{1.05}
\begin{tabular}{@{}>{\raggedright\arraybackslash}p{.10\linewidth}
                  >{\raggedright\arraybackslash}p{.28\linewidth}
                  >{\raggedright\arraybackslash}p{.28\linewidth}
                  >{\raggedright\arraybackslash}p{.28\linewidth}@{}}
\toprule
\textbf{Method} & \textbf{Runtime information} & \textbf{Model behind the guarantee}
  & \textbf{Dependencies between agents} \\
\midrule
\cite{ElSayedAlyBAETF21}
& joint state; communication with the shields
& finite joint abstraction
& centralised, or factored with coordination \\
\midrule
\cite{MelcerAT22}
& shared abstract label and own action; internal shield state
& centralised shield over a finite abstraction
& inherited from the centralised shield \\
\midrule
\cite{MelcerAT24}
& own observations, bounded history
& centralised shield over a finite joint abstraction
& inherited from the centralised shield \\
\midrule
\cite{YangY26}
& own observations, as beliefs about a global machine
& global Mealy machine
& projected from the global machine \\
\midrule
\cite{BrorholtLS25}
& current projection, no memory
& finite state space, projected per agent
& acyclic: ordered assume-guarantee \\
\midrule
\cite{AdalatHB26}
& global state, or contract and local history if they fix the mask; contract selection synchronised in training
& finite global model; contracts certified together
& circular, certified together \\
\midrule
This work
& own actions and observations, as beliefs
& per-agent model with public types; finite reachable beliefs for synthesis, infinite worlds for guards
& mutual on current safety and action bounds; acyclic on safety after the step \\
\bottomrule
\end{tabular}
\caption{Multi-agent shielding methods compared with epistemic typestates.}
\label{tab:positioning}
\end{table}
General distributed controller synthesis under differing observations is undecidable~\cite{PnueliR90}, and all the
shields in Table~\ref{tab:positioning} are instead derived from a model of the whole team. Whether an
agent can run its part from its own information is then a property of the derived shield.

An epistemic typestate is built for one agent, from its own world dynamics, its observations and the public types of
the other agents. No joint state space is required. Public types act as interfaces: an agent relies only on what the others'
public types permit and guarantee. The agent runs its typestate on its
own actions and observations, so execution without communication holds by construction. Declarations let an agent
commit to actions that others can rely on. Unlike the contracts of Adalat et al.~\cite{AdalatHB26}, they are checked
with the agent's own typestate. Team safety then follows from these separate checks (Section~\ref{sec:teams}):
agents may rely on each other's current safety and public action bounds, and only reliance on safety after the step
must be acyclic, as in circular assume-guarantee reasoning~\cite{MisraC81,AbadiL95}. None of the
compared methods builds each agent's safety mechanism separately while allowing mutual reliance. 
Rather than a model of the whole team, a typestate requires finitely many reachable beliefs of its agent, with the computability conditions of
Section~\ref{sec:building}, or a guard supplied by the designer, which also covers infinite world spaces. A typestate uses only what its agent can know: where a shield reads state that the
agent cannot sense, the typestate may permit fewer actions. Section~\ref{sec:evaluation} compares the two
quantitatively with Adalat et al.~\cite{AdalatHB26} and Brorholt et al.~\cite{BrorholtLS25}.

\section{Epistemic typestates}\label{sec:ets}

We now introduce epistemic typestates and the theoretical machinery supporting safe learning inside them.
 
\begin{definition}[Worlds]\label{def:worlds}
Let $V$ be a set of \emph{variables} and $D$ a set of \emph{values}. A \emph{world} is a partial function
$w:V\rightharpoonup D$. $W$ is the set of all worlds.
\end{definition}
A world is a snapshot of the values of the variables of a system, observed and hidden. In our running example, the worlds of car~1 assign
its speed $v$, the speed $v_f$ of car~2, its gap $d$, the speed $v_{ff}$ of car~3 and the gap $d_f$
of car~2, with speeds in $\{0,2\}$ and gaps in $\mathbb{Z}$. We write them as $(v,v_f,d,v_{ff},d_f)$.

\begin{definition}[World dynamics]\label{def:dyn}
Let $A$ be a finite nonempty set of \emph{actions}, those the agent can physically take. The \emph{world dynamics} are given by a
function $\Post:W\times A\to 2^{W}$: $\Post(w,a)$ is the set of worlds that may follow $w$ when the agent takes action
$a$.
\end{definition}
The world dynamics describe how the environment evolves around it. $\Post$ is
nondeterministic because it accounts for every behaviour of the environment and the peers that the agent cannot rule
out.

In our running example, car~1 has
$A=\{\mathit{brake},\mathit{coast},\mathit{accelerate}\}$. Speeds change by $-2$, $0$ or $+2$ within $[0,2]$, and gaps follow
\[ d'=d+\big((v_f-v)+(v'_f-v')\big)/2, \]
where primes denote values after the step. Car~2 may take any action
permitted by its public type (here that of its typestate synthesised without declarations), and the successors keep
car~2's gap strictly between $0$ and $20$, the safety that car~2 guarantees. For instance, $\Post((0,0,10,\allowbreak 0,10),\mathit{coast})$
has four worlds: car~2 and car~3 may each stay still or accelerate.

\begin{definition}[Observation function]\label{def:obs}
Let $V_o\subseteq V$ be the \emph{observed} variables. The \emph{observation function} $\Obs:W\to W$ is
\[ \Obs(w)=\{x\mapsto w(x)\mid x\in\mathrm{dom}(w)\cap V_o\}. \]
Two worlds $w,w'$ are \emph{indistinguishable} if $\Obs(w)=\Obs(w')$.
\end{definition}
In our running example, car~1 observes
$V_o=\{v,v_f,d\}$, the quantities its sensors read. We write its observations as tuples $(v,v_f,d)$. The two configurations of
Section~\ref{sec:running} are indistinguishable:
\[ \Obs(0,2,11,2,10)=\Obs(0,2,11,0,9)=(0,2,11). \]

\begin{definition}[Obligations and restrictions]\label{def:obl}
A \emph{trigger} is a predicate $\tau$ over worlds:
$\tau(w)=\tau(w')$ for all $w,w'\in W$ with $\Obs(w)=\Obs(w')$. An \emph{obligation} is a pair $(\tau,a)$ of a trigger and an
action $a\in A$, and a \emph{restriction} is a pair $(\tau,R)$ of a trigger and a set of actions $R\subseteq A$. The
\emph{declarations} $C$ are a finite sequence of obligations and restrictions.
\end{definition}
An obligation $(\tau,a)$ reads ``when $\tau$ is triggered, take $a$ if it is admissible''
(Definition~\ref{def:adm}), and a restriction $(\tau,R)$ reads ``when $\tau$
is triggered, take only actions in $R$''. Triggers depend only on
what the agent observes, so the agent can always evaluate them. What the agent remembers of its own past, e.g., its last
action, can be held by observed variables that $\Post$ updates, so triggers may read it. In our running example, \emph{car~1} may declare
$C=\big((v\geq 2,\ \mathit{brake}),\ (d\leq 3,\ \{\mathit{brake},\mathit{coast}\})\big)$: ``at top speed, brake'' and
``with a gap of at most $3$, never accelerate''. Here $v\geq 2$ is the predicate that holds in $w$ if $w$ assigns $v$
and $w(v)\geq 2$, and similarly for $d\leq 3$.

\begin{definition}[Belief update]\label{def:update}
The \emph{update} of a set of worlds $K\subseteq W$ by action $a\in A$ and observation $u$ is
\[ K\cdot(a,u)=\{w'\in\Post(w,a)\mid w\in K,\ \Obs(w')=u\}. \]
\end{definition}
Belief updates yield the set of worlds that may follow a world of $K$ under $a$ and show the observation $u$. In our running example, the four successors above split by
observation into two sets:
\begin{align*}
\{w_0\}\cdot(\mathit{coast},(0,2,11))&=\{(0,2,11,0,9),\ (0,2,11,2,10)\},\\
\{w_0\}\cdot(\mathit{coast},(0,0,10))&=\{(0,0,10,0,10),\ (0,0,10,2,11)\},
\end{align*}
with $w_0=(0,0,10,0,10)$. After coasting and reading $(0,2,11)$, car~1 knows that car~2
accelerated, but not whether car~3 did.

\subsection{Epistemic typestates and safety by construction}\label{sec:typestates}

An epistemic typestate records the current belief and specifies which actions the agent may take. Its transitions
use only the action taken and the next observation.

\begin{definition}[Epistemic typestate]\label{def:ets}
Let $A$, $V_o$ and $C$ be actions, observed variables and declarations (Definitions~\ref{def:dyn}, \ref{def:obs}
and~\ref{def:obl}), let the \emph{safety requirement} $\safe$ be a predicate over worlds, with $w$ \emph{safe} if
$\safe(w)$, and let $I\subseteq W$ be the \emph{initial worlds}, a nonempty set of
indistinguishable safe worlds where the agent may start. An \emph{epistemic typestate} is a tuple $T=(Q,\allowbreak A,\allowbreak V_o,\allowbreak I,\allowbreak q_0,\allowbreak \kn,\allowbreak \delta,\allowbreak \may,\allowbreak \safe,\allowbreak C)$ where
\begin{itemize}
\item $Q$ is a set of \emph{states};
\item $q_0\in Q$ is the initial state;
\item $\kn:Q\to 2^{W}\setminus\{\emptyset\}$ maps each state $q$ to its \emph{belief} $\kn(q)$, a set of
  indistinguishable worlds: $\Obs(w)=\Obs(w')$ for all $w,w'\in\kn(q)$. The initial state believes the initial
  worlds: $\kn(q_0)=I$;
\item $\delta:Q\times A\times W\rightharpoonup Q$ is the \emph{transition} function, which updates the belief: for every
  $q\in Q$, $a\in\may(q)$ and $w\in W$, $\delta(q,a,\Obs(w))$ is defined iff
  $\kn(q)\cdot(a,\Obs(w))\neq\emptyset$, and then
  \[ \kn(\delta(q,a,\Obs(w)))=\kn(q)\cdot(a,\Obs(w)), \]
  and $\delta$ is undefined elsewhere;
\item $\may:Q\to 2^{A}$ gives the \emph{permitted} actions.
\end{itemize}
Let $\mathcal{T}$ be the set of epistemic typestates.
\end{definition}

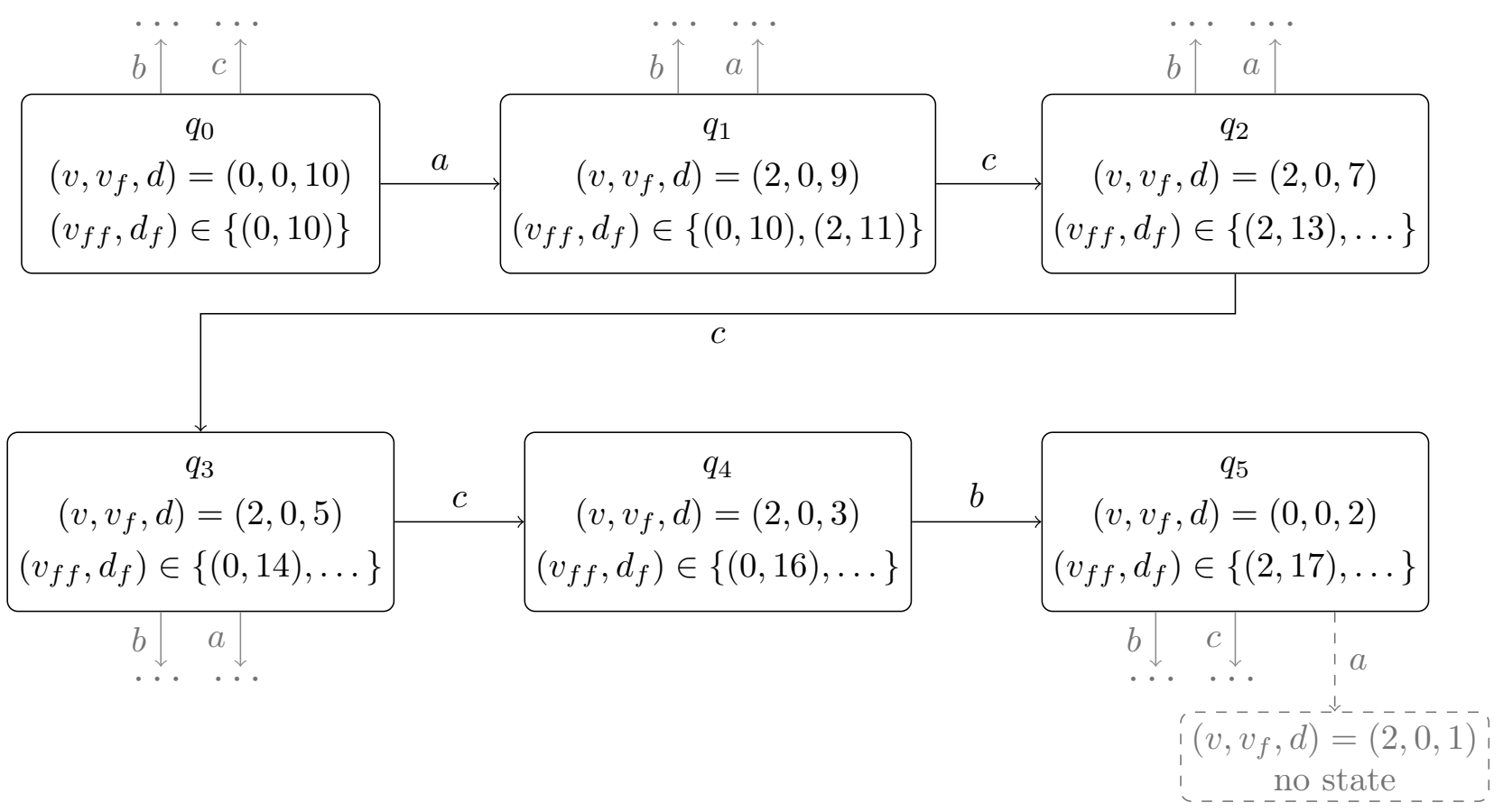
\begin{figure}[H]
\centering
\newcommand{\tsnode}[5]{\node[ts] (#1) at (#2) {%
  $#3$\\[3pt]
  $(v,v_f,d)=#4$\\[3pt]
  $(v_{ff},d_f)\in#5$};}
\newcommand{\up}[3]{\draw[ghost] ([xshift=#2]#1.north) -- node[lab, left] {$#3$} ++(0,0.55)
  node[above, inner sep=1pt] {$\cdots$};}
\newcommand{\down}[3]{\draw[ghost] ([xshift=#2]#1.south) -- node[lab, left] {$#3$} ++(0,-0.55)
  node[below, inner sep=1pt] {$\cdots$};}
\resizebox{0.8\linewidth}{!}{%
\begin{tikzpicture}[ts/.style={draw, rounded corners=3pt, align=center, font=\footnotesize, inner sep=3pt,
    minimum width=3.6cm, minimum height=1.8cm}, lab/.style={font=\footnotesize},
    ghost/.style={->, black!45, text=black!55}]
  \tsnode{q0}{0,0}{q_0}{(0,0,10)}{\{(0,10)\}}
  \tsnode{q1}{5.2,0}{q_1}{(2,0,9)}{\{(0,10),(2,11)\}}
  \tsnode{q2}{10.4,0}{q_2}{(2,0,7)}{\{(2,13),\dots\}}
  \tsnode{q3}{0,-3.4}{q_3}{(2,0,5)}{\{(0,14),\dots\}}
  \tsnode{q4}{5.2,-3.4}{q_4}{(2,0,3)}{\{(0,16),\dots\}}
  \tsnode{q5}{10.4,-3.4}{q_5}{(0,0,2)}{\{(2,17),\dots\}}
  \draw[->] (q0) -- node[lab, above] {$a$} (q1);
  \draw[->] (q1) -- node[lab, above] {$c$} (q2);
  \draw[->] (q2.south) -- ++(0,-0.4) -| node[lab, pos=0.25, below] {$c$} (q3.north);
  \draw[->] (q3) -- node[lab, above] {$c$} (q4);
  \draw[->] (q4) -- node[lab, above] {$b$} (q5);
  \up{q0}{-4mm}{b} \up{q0}{4mm}{c}
  \up{q1}{-4mm}{b} \up{q1}{4mm}{a}
  \up{q2}{-4mm}{b} \up{q2}{4mm}{a}
  \down{q3}{-4mm}{b} \down{q3}{4mm}{a}
  \down{q5}{-8mm}{b} \down{q5}{0mm}{c}
  \draw[->, dashed, black!55] ([xshift=10mm]q5.south) -- node[lab, right] {$a$} ++(0,-1.0)
    node[draw, dashed, rounded corners=3pt, align=center, font=\footnotesize, inner sep=3pt, anchor=north]
    {$(v,v_f,d)=(2,0,1)$\\ no state};
\end{tikzpicture}
}
\caption{A path through car~1's typestate, with $C$ empty. Boxes show the observation and the possible hidden pairs,
with dots for further pairs; $b$, $c$ and $a$ mean brake, coast and accelerate. Black arrows trace the path, grey
arrows other permitted actions, and the dashed arrow an excluded one; other successors are omitted.}
\label{fig:path}
\end{figure}

In our running example, Figure~\ref{fig:path} shows a path through car~1's epistemic typestate. Each box records what car~1
observes and the hidden configurations it still considers possible. Car~1 moves between states according to
its action and the next observation. Along the black path, car~2 stays at rest while uncertainty about
car~3 grows. The outgoing solid arrows give the learner its choices. The typestate may
forbid acceleration that is harmless immediately but could make a later collision unavoidable.

\begin{definition}[Admissible actions]\label{def:adm}
Let $Q$ be a set of states with beliefs $\kn$. The \emph{admissible} actions at
a state $q\in Q$ are
\[
\begin{aligned}
\mathit{adm}(q)=\{\,a\in A\mid{}&\forall w\in\kn(q).\ \forall w'\in\Post(w,a).\ \safe(w')\ \\
 &\wedge\ \exists q'\in Q.\ \kn(q')=\kn(q)\cdot(a,\Obs(w'))\,\}.
\end{aligned}
\]
\end{definition}

\begin{definition}[Obligatory action]\label{def:bind}
Let $Q$ be a set of states with beliefs $\kn$, and $C=(c_1,\dots,c_n)$
declarations. For a state
$q\in Q$, the indices of the obligations that are triggered at $q$ and whose action is admissible are
\[ J(q)=\{\,1\leq i\leq n\mid c_i=(\tau_i,a_i)\ \wedge\ \big(\forall w\in\kn(q).\ \tau_i(w)\big)\ \wedge\
   a_i\in\mathit{adm}(q)\,\}. \]
The \emph{obligatory action} is the partial function $\must:Q\rightharpoonup A$ with
$\mathrm{dom}(\must)=\{q\in Q\mid J(q)\neq\emptyset\}$ and $\must(q)=a_x$, with $x=\min J(q)$.
\end{definition}

\begin{definition}[Actions permitted in a world]\label{def:world-actions}
Let $T=(Q,\allowbreak A,\allowbreak V_o,\allowbreak I,\allowbreak q_0,\allowbreak \kn,\allowbreak \delta,\allowbreak \may,\allowbreak \safe,\allowbreak C) \in \mathcal{T}$. The function $g_T:W\to 2^{A}$ gives the actions
permitted in each world:
\[ g_T(w)=\bigcup_{q\in Q}\{\,a\in\may(q)\mid \exists w'\in\kn(q).\ \Obs(w')=\Obs(w)\,\}. \]
\end{definition}
The set $g_T(w)$ collects the actions permitted at all states with observation $(v,v_f,d)$. It
describes what car~1 might do without revealing its current belief.

\begin{definition}[Well-formedness]\label{def:wf}
Let $T=(Q,\allowbreak A,\allowbreak V_o,\allowbreak I,\allowbreak q_0,\allowbreak \kn,\allowbreak \delta,\allowbreak \may,\allowbreak \safe,\allowbreak C)\in\mathcal{T}$. A state $q\in Q$ is \emph{well typed} in $T$, and $T$ is \emph{well formed}, by the rules
\begin{mathpar}
\inferrule*[right=State]
  {\may(q)\subseteq\mathit{adm}(q) \\
   \may(q)\neq\emptyset \\\\
   \forall(\tau,R)\in C.\ \ \big(\forall w\in\kn(q).\ \tau(w)\big)\Rightarrow\may(q)\subseteq R \\\\
   q\in\mathrm{dom}(\must)\Rightarrow\may(q)=\{\must(q)\}}
  {(Q,\allowbreak A,\allowbreak V_o,\allowbreak I,\allowbreak q_0,\allowbreak \kn,\allowbreak \delta,\allowbreak \may,\allowbreak \safe,\allowbreak C)\vdash q}
\\
\inferrule*[right=Type]
  {\forall q\in Q.\ \ (Q,\allowbreak A,\allowbreak V_o,\allowbreak I,\allowbreak q_0,\allowbreak \kn,\allowbreak \delta,\allowbreak \may,\allowbreak \safe,\allowbreak C)\vdash q}
  {\vdash(Q,\allowbreak A,\allowbreak V_o,\allowbreak I,\allowbreak q_0,\allowbreak \kn,\allowbreak \delta,\allowbreak \may,\allowbreak \safe,\allowbreak C)}
\end{mathpar}
\end{definition}

The State rule leaves at least one action, permits only actions that are safe in every possible world and have
a continuation, and enforces the declarations. Triggered restrictions narrow the choices (the first triggered
obligation whose action is admissible fixes the choice). An obligation cannot be evaded by omitting its action
from the permitted set. The Type rule requires these checks at every state. They concern the typestate, not the
learner: the learner chooses among the actions its typestate permits, so every policy it learns conforms by
construction.

\begin{definition}[Public type]\label{def:public}
Let  $T=(Q,\allowbreak A,\allowbreak V_o,\allowbreak I,\allowbreak q_0,\allowbreak \kn,\allowbreak \delta,\allowbreak \may,\allowbreak \safe,\allowbreak C)\in\mathcal{T}$, 
and $g_T$ yielding the actions permitted in a world. 
A \emph{public type} of $T$ is a tuple $(V_o,\rho,\safe)$ with $\rho:W\to 2^{A}$ such that $g_T(w)\subseteq\rho(w)$
for every $w\in W$. The most precise one is $\mathit{Pub}(T)=(V_o,g_T,\safe)$.
\end{definition}

A public type summarises the behaviour of an agent following a well-formed typestate. It bounds the actions the
agent may take and states its safety guarantee, while its current belief and learning policy remain private. Other
agents use this information to account for its possible behaviour when deciding what is safe for themselves. Any other public type over-approximates it and can be written by the designer
in advance, e.g., when agents rely on each other and neither typestate can be built first. The condition
$g_T\subseteq\rho$ then says that the agent keeps its promise.
In our running example, car~1 uses car~2's public type to account for what car~2 may do under the hidden
configurations in car~1's belief.
The guarantee concerns every policy that chooses from the permitted actions.

\begin{definition}[Run]\label{def:run}
Let $T=(Q,\allowbreak A,\allowbreak V_o,\allowbreak I,\allowbreak q_0,\allowbreak \kn,\allowbreak \delta,\allowbreak \may,\allowbreak \safe,\allowbreak C)\in\mathcal{T}$. A \emph{run} of $T$ is a sequence
$(w_t,q_t,a_t)_{t\in\mathbb{N}}$ of worlds, states and actions with $w_0\in I$ and,
for every $t\in\mathbb{N}$, $a_t\in\may(q_t)$, $w_{t+1}\in\Post(w_t,a_t)$ and $q_{t+1}=\delta(q_t,a_t,\Obs(w_{t+1}))$.
\end{definition}
A run records the actual car configurations together with car~1's typestate states and
chosen actions. If the learner accelerates at $q_0$ and cars~2 and~3 remain at rest, the world changes from
$(0,0,10,0,10)$ to $(2,0,9,0,10)$. The observation $(2,0,9)$ leads to $q_1$ in Figure~\ref{fig:path}, whose belief
also includes a world where car~3 accelerated.

\begin{theorem}[Type soundness]\label{thm:sound}
Let $T=(Q,\allowbreak A,\allowbreak V_o,\allowbreak I,\allowbreak q_0,\allowbreak \kn,\allowbreak \delta,\allowbreak \may,\allowbreak \safe,\allowbreak C)\in\mathcal{T}$ with $\vdash T$.
\begin{itemize}
\item \emph{Progress}: for every $q\in Q$, $w\in\kn(q)$, $a\in\may(q)$ and $w'\in\Post(w,a)$, $\delta(q,a,\Obs(w'))$ is
  defined and $w'\in\kn(\delta(q,a,\Obs(w')))$.
\item \emph{Preservation and safety}: in every run of $T$, for every $t\in\mathbb{N}$, $w_t\in\kn(q_t)$,
  $\safe(w_t)$ and $\may(q_t)\neq\emptyset$.
\end{itemize}
\end{theorem}
Progress means that the typestate can follow every possible outcome of a permitted action. If $\Post(w,a)$ is empty,
it is the world dynamics that have no continuation, not the agent. Preservation means
that the belief continues to include the actual world. Throughout every run, the safety requirement holds and
at least one action remains permitted. These guarantees hold for any policy that chooses permitted actions,
provided the environment follows the specified world dynamics.

\subsection{Building epistemic typestates}\label{sec:building}

Well-formed typestates can be synthesised from the model or obtained from a guard that selects actions from the
current observation while preserving a set of safe worlds. Synthesis starts with beliefs reachable through actions
that are safe for one step, then repeatedly removes those with no permitted action. This is the knowledge-based
subset construction for safety games with imperfect information~\cite{Reif84,ChatterjeeDHR07}, applied to a single
agent: the possible successors come from the environment and from the other agents' public types, and declarations
constrain the agent's own choices.
\begin{definition}[Reachable states]\label{def:reachstates}
Let $T=(Q,\allowbreak A,\allowbreak V_o,\allowbreak I,\allowbreak q_0,\allowbreak \kn,\allowbreak \delta,\allowbreak \may,\allowbreak \safe,\allowbreak C)\in\mathcal{T}$. The \emph{reachable states}
$\mathit{Reach}(T)$ are the least $S\subseteq Q$ such that
\[
\begin{gathered}
q_0\in S\ \wedge\ \Big(\forall q\in S.\ \forall a\in A.\ \forall w\in W.\ \big((q,a,\Obs(w))\in\mathrm{dom}(\delta)\\
\Rightarrow\delta(q,a,\Obs(w))\in S\big)\Big).
\end{gathered}
\]
\end{definition}

Reachable states are those the agent can visit from the initial one.
The states $q_0,\dots,q_5$ along the path in Figure~\ref{fig:path} are reachable.

\begin{definition}[Reachable beliefs]\label{def:reach}
The \emph{reachable beliefs} $K_r$ are such that $I\in K_r$ and, for every $K\in K_r$ and $a\in A$,
\[
\begin{gathered}
\big(\forall w\in K.\ \forall w'\in\Post(w,a).\ \safe(w')\big)\ \Rightarrow\\
\forall u\in K.\ \forall u'\in\Post(u,a).\ K\cdot(a,\Obs(u'))\in K_r.
\end{gathered}
\]
\end{definition}
These are candidate beliefs obtained from $I$ by belief updates along actions that are safe for one step.
Reaching a belief safely does not guarantee that safety can be maintained from it. We consider only finite $K_r$.

\begin{assumption}[Consistent declarations]\label{ass:cons}
The declarations are consistent:
\[ \forall(\tau,a),(\tau',R)\in C.\ \forall K\in K_r.\ \forall w\in K.\ \big(\tau(w)\wedge\tau'(w)\big)\Rightarrow a\in R. \]
\end{assumption}
An obligation must not require an action that a simultaneously triggered restriction forbids. We assume this
consistency throughout. %

\begin{definition}[Permitted actions]\label{def:perm}
Let $X\subseteq K_r$ and $Q=\{q_K\mid K\in X\}$, where $q_K$ is a state with belief $\kn(q_K)=K$, and
$q_K=q_{K'}\Rightarrow K=K'$ for all $K,K'\in X$. For $K\in X$,
\begin{align*}
\mathit{perm}(X,K)=\{\,a\in\mathit{adm}(q_K)\mid\ & \big(\forall(\tau,R)\in C.\ (\forall w\in K.\ \tau(w))\Rightarrow
  a\in R\big)\ \wedge\\
  & \big(q_K\in\mathrm{dom}(\must)\Rightarrow a=\must(q_K)\big)\,\}.
\end{align*}
\end{definition}
An action is available only if it is safe and every observation leads to a belief in the current
candidate set. Declarations then narrow these choices.

\begin{definition}[Pruning]\label{def:prune}
The \emph{pruned beliefs} are
\[ X^\ast=\bigcap_{i\in\mathbb{N}}X_i, \]
where $X_0=K_r$ and $X_{i+1}=\{K\in X_i\mid \mathit{perm}(X_i,K)\neq\emptyset\}$ for every $i\in\mathbb{N}$.
\end{definition}

\begin{lemma}[Stabilisation]\label{lem:stab}
If $K_r$ is finite, there is $n\in\mathbb{N}$ with $X_{n+1}=X_n$, and $X^\ast=X_n$.
\end{lemma}

Each round removes beliefs with no permitted action. Their removal can make actions at other beliefs inadmissible,
so the test repeats. %
In our running example, for car~1 with $C$ empty, pruning removes the beliefs with observation $(2,0,1)$, where every
action may lead to a collision, and those with $(0,2,19)$, where every action may let the gap reach $20$.

\begin{definition}[Synthesised typestate]\label{def:synth}
For all $K, K' \subseteq W$, let $q_K$ be a state with $q_K=q_{K'} \Rightarrow K=K'$. If
$I\in X^\ast$, the \emph{synthesised typestate} is
\[ T^\ast=(Q^\ast,\allowbreak A,\allowbreak V_o,\allowbreak I,\allowbreak q^\ast_0,\allowbreak \kn^\ast,\allowbreak \delta^\ast,\allowbreak \may^\ast,\allowbreak \safe,\allowbreak C) \]
where
\begin{itemize}
\item $Q^\ast=\{q_K\mid K\in X^\ast\}$;
\item $q^\ast_0=q_I$, with $\kn^\ast(q_I)=I$;
\item $\kn^\ast(q_K)=K$ for every $q_K\in Q^\ast$;
\item $\delta^\ast$ is the partial function with $\delta^\ast(q_K,a,w')=q_{K\cdot(a,w')}$ for every $q_K\in Q^\ast$,
  $a\in\may^\ast(q_K)$ and $w'\in W$ with $K\cdot(a,w')\neq\emptyset$, and undefined elsewhere;
\item $\may^\ast(q_K)=\mathit{perm}(X^\ast,K)$ for every $q_K\in Q^\ast$.
\end{itemize}
Let $\mathcal{T}^\ast$ be the set of synthesised typestates.
\end{definition}

In our running example with $C$ empty, the surviving beliefs become the states of car~1's typestate.
Figure~\ref{fig:path} shows some of them. The initial belief gives $q_0$. Acceleration followed by observation
$(2,0,9)$ updates this belief to the one at $q_1$, giving the first black transition. At $q_5$, braking and
coasting remain permitted, but acceleration is excluded because it can produce observation $(2,0,1)$,
leading to a belief removed by pruning.

\begin{proposition}[Synthesis soundness]\label{prop:wf}
If $T^\ast\in\mathcal{T}^\ast$, then $T^\ast\in\mathcal{T}$ and $\vdash T^\ast$.
\end{proposition}

\begin{proposition}[Synthesis completeness]\label{prop:complete}
$(\exists T\in\mathcal{T}.\ \vdash T)\iff\mathcal{T}^\ast\neq\emptyset$.
\end{proposition}

Soundness means that whenever synthesis succeeds, it produces a well-formed typestate, so any policy following it
satisfies the safety requirement. Completeness means that synthesis succeeds whenever any well-formed typestate
exists for the same model and declarations, including handwritten ones. 

\begin{proposition}[Maximal permissiveness]\label{prop:greatest}
Let $T=(Q,\allowbreak A,\allowbreak V_o,\allowbreak I,\allowbreak q_0,\allowbreak \kn,\allowbreak \delta,\allowbreak \may,\allowbreak \safe,\allowbreak C)\in\mathcal{T}$ with $\vdash T$, and
$T^\ast=(Q^\ast,\allowbreak A,\allowbreak V_o,\allowbreak I,\allowbreak q^\ast_0,\allowbreak \kn^\ast,\allowbreak \delta^\ast,\allowbreak \may^\ast,\allowbreak \safe,\allowbreak C)\in\mathcal{T}^\ast$, with $\must^\ast$ the
obligatory action for $Q^\ast$ and $\kn^\ast$. Then
\[
\begin{gathered}
\forall q\in\mathit{Reach}(T).\ \exists q^\ast\in Q^\ast.\ \big(\kn^\ast(q^\ast)=\kn(q)\ \wedge\\
(q^\ast\notin\mathrm{dom}(\must^\ast)\Rightarrow\may(q)\subseteq\may^\ast(q^\ast))\big).
\end{gathered}
\]
\end{proposition}

Without obligations, synthesis leaves the learner as much freedom as any well-formed typestate could offer during
execution, given the same belief. Obligations qualify this guarantee: synthesis may have to require a particular action
even when another typestate would allow alternatives.

When $I$ and every $\Post(w,a)$ are finite and computable, and world equality, $\Obs$, $\safe$ and triggers are
computable, every reachable belief is a finite set of worlds, and $K_r$, every $\mathit{perm}(X,K)$ and, by
Lemma~\ref{lem:stab}, $X^\ast$ can be computed.

\begin{proposition}[Typestates from guards]\label{prop:guard}
Let $C$ be empty, $I\subseteq Y\subseteq W$, and $G:W\to 2^{A}$ a \emph{guard} with $G(w)=G(w')$ for all
$w,w'\in W$ with $\Obs(w)=\Obs(w')$. Suppose that, for every $w\in Y$, $\safe(w)$, $G(w)\neq\emptyset$ and
$\Post(w,a)\subseteq Y$ for every $a\in G(w)$. Then, there is
$T_G=(Q,\allowbreak A,\allowbreak V_o,\allowbreak I,\allowbreak q_0,\allowbreak \kn,\allowbreak \delta,\allowbreak \may,\allowbreak \safe,\allowbreak C)\in\mathcal{T}$ with $\vdash T_G$ and $\may(q)=G(w)$ for every $q\in Q$
and $w\in\kn(q)$. Hence, $g_{T_G}(w)\subseteq G(w)$ for every $w\in W$.
\end{proposition}

A guard reads the current observation, which, as for triggers, can hold what the agent remembers of its past. It
neither enumerates nor tracks beliefs. Within a safe region containing
all possible starts, it must always offer an action, and every offered action must keep the agent there. This
gives a well-formed typestate and public action bounds even when $K_r$ is infinite. %

\subsection{Team safety from local knowledge}\label{sec:teams}

We now relate each agent's world dynamics to what happens when agents act together. In a team of a finite set $N$
of agents, each agent $i\in N$ has actions $A_i$, observation function $\Obs_i$, safety requirement $\safe_i$, initial
worlds $I_i$, world dynamics $\Post_i$, a typestate $T_i$ with permitted actions $\may_i$, beliefs $\kn_i$ and
transitions $\delta_i$, a public type $(V_{o,i},\rho_i,\safe_i)$ of $T_i$, and a set $\mathit{Rel}_i\subseteq N$ of
agents whose safety after the step its world dynamics assume.

\begin{definition}[Team run]\label{def:team}
A \emph{team run} is a family of sequences $(w_{i,t},\allowbreak q_{i,t},\allowbreak a_{i,t})_{t\in\mathbb{N}}$, one for every $i\in N$, with
$w_{i,0}\in I_i$, $q_{i,0}$ the initial state of $T_i$, and, for every $t\in\mathbb{N}$, $a_{i,t}\in\may_i(q_{i,t})$ and
$q_{i,t+1}=\delta_i(q_{i,t},a_{i,t},\Obs_i(w_{i,t+1}))$.
\end{definition}
Each agent chooses a permitted action and updates its typestate from its own next observation. The worlds evolve
as the environment and the other agents act. In our running example, $w_{i,t}$ is the true world of follower $i$ at time $t$, e.g., $(v,v_f,d,v_{ff},d_f)$
for car~1. Car~1's $d_f$ and the gap $d$ of car~2 are the same quantity. Actors that are not agents, such as the
lead car, affect how the worlds evolve. The following theorem gives conditions under which these actual successors
belong to each agent's world dynamics.

\begin{definition}[Acyclic safety reliance]\label{def:reliance}
The \emph{safety reliance} $\{(i,j)\in N\times N\mid j\in \mathit{Rel}_i\}$ is \emph{acyclic} if its transitive
closure contains no pair $(i,i)$.
\end{definition}

In our running example, car~1 relies on car~2's safety after the step, $\mathit{Rel}_{\text{car~1}}=\{\text{car~2}\}$. Each car relies only
on cars ahead of it, so following these dependencies never returns to the same car.

\begin{theorem}[Team safety]\label{thm:safety}
Let $\vdash T_i$ for every $i\in N$, let safety reliance be acyclic, and let $H_t$ hold if, for every $i\in N$,
\[
\begin{gathered}
\big(\forall j\in N.\ a_{j,t}\in\rho_j(w_{j,t})\wedge\safe_j(w_{j,t})\big)\wedge\big(\forall j\in \mathit{Rel}_i.\ \safe_j(w_{j,t+1})\big)\\
\Rightarrow w_{i,t+1}\in\Post_i(w_{i,t},a_{i,t}).
\end{gathered}
\]
\begin{itemize}
\item[(a)] In every team run such that $H_t$ holds for every $t\in\mathbb{N}$, for every agent $i\in N$ the sequence
  $(w_{i,t},\allowbreak q_{i,t},\allowbreak a_{i,t})_{t\in\mathbb{N}}$ is a run of $T_i$.
\item[(b)] Let $t\in\mathbb{N}$ and, for every $i\in N$, let $(w_{i,s})_{s\leq t+1}$, $(q_{i,s})_{s\leq t}$ and
  $(a_{i,s})_{s\leq t}$ be worlds, states and actions with $w_{i,0}\in I_i$, $q_{i,0}$ the initial state of $T_i$,
  $a_{i,s}\in\may_i(q_{i,s})$ for every $s\leq t$, and $q_{i,s+1}=\delta_i(q_{i,s},a_{i,s},\Obs_i(w_{i,s+1}))$ for every
  $s<t$. If $H_s$ holds for every $s\leq t$, then $\delta_i(q_{i,t},\allowbreak a_{i,t},\allowbreak \Obs_i(w_{i,t+1}))$
  is defined for every $i\in N$.
\end{itemize}
\end{theorem}
The theorem combines the agents' local safety guarantees. Each model must cover what can happen when all agents
are currently safe and choose actions within their public types, as their typestates ensure, provided the agents it relies on remain safe
after the step. If this reliance is acyclic, every actual successor is among the worlds its agent's model allows.
Each typestate therefore keeps its agent safe (Theorem~\ref{thm:sound}), for any policy choosing permitted actions
and without communication. By (b), while $H_t$ holds, every typestate update is defined, so no typestate gets stuck.
Agents may rely mutually on their current safety and on the action bounds of their public types. Only reliance on
safety after the step must be acyclic. Each
agent can therefore be checked separately against the others' public types. Public types thus play the role of
interfaces~\cite{AlfaroH01}, and building each typestate against the others' public types is a form of
assume-guarantee synthesis~\cite{ChatterjeeH07}, here for one agent at a time.
In our running example, the gap update describes how the cars actually move. Car~1's model must account for every
action allowed by car~2's public type and every action of car~3.

\subsection{Shields as a special case}\label{sec:shields}

Shields that guarantee safety in every run differ in what they read and in who chooses the action, but they share a
formal core: a set of worlds from which safety can be maintained, and in each of them a nonempty set of actions that
keep the system there. We express this core in our notation and show that it corresponds to epistemic typestates with
full observation and no declarations. Full observation here is observation of the world represented by the shield. That world may itself
be an abstraction or a local projection.

\begin{definition}[Shield]\label{def:shield}
A \emph{shield} is a pair $(Z,\sigma)$ of a set $Z\subseteq W$ of worlds and a function $\sigma:Z\to 2^{A}$ such that,
for every $z\in Z$, $\safe(z)$, $\sigma(z)\neq\emptyset$, and $\Post(z,a)\subseteq Z$ for every $a\in\sigma(z)$. The
\emph{winning worlds} are
\[ \mathit{Win}=\bigcup\{\,Z\subseteq W\mid \forall w\in Z.\ \safe(w)\wedge\exists a\in A.\ \Post(w,a)\subseteq Z\,\}, \]
and $\mathit{shield}(w)=\{\,a\in A\mid \Post(w,a)\subseteq\mathit{Win}\,\}$ for every $w\in\mathit{Win}$.
\end{definition}
Following a shield keeps the agent within $Z$, hence safe, whatever outcome its permitted actions produce. Here a
world represents the state the shield reads: an abstract environmental state~\cite{BloemKKW15,AlshiekhBEKNT18}, an
observed projection of the global state~\cite{BrorholtLS25}, or the global state itself~\cite{AdalatHB26}.
Requirements given by safety automata fit by adding the automaton's current state to each world, and $\safe$ then
checks that no violation has been recorded. The nondeterminism of $\Post$ accounts for the choices of the
environment, including other agents. 
This view also covers shields that correct chosen actions: $\sigma$ describes their possible outputs (for a
centralised shield, these are joint actions). %

\begin{lemma}[Maximally permissive shield]\label{lem:maxshield}
$(\mathit{Win},\mathit{shield})$ is a shield, and every shield $(Z,\sigma)$ satisfies $Z\subseteq\mathit{Win}$ and
$\sigma(z)\subseteq\mathit{shield}(z)$ for every $z\in Z$.
\end{lemma}

$\mathit{Win}$ contains exactly the worlds from which the agent can always remain safe, whatever the environment
does~\cite{BernetJW02}. The shield permits every action that keeps all possible outcomes within this region. It
therefore gives the agent the greatest possible freedom while guaranteeing safety: this is the maximally
permissive pre-emptive shield~\cite{AlshiekhBEKNT18}.

\Needspace{10\baselineskip}
\begin{theorem}[Shields are epistemic typestates]\label{thm:shields}
Let $V_o=V$, $C$ be empty and $I=\{w_0\}$.
\begin{itemize}
\item[(a)] Let $(Z,\sigma)$ be a shield with $w_0\in Z$, and, for every $z\in Z$, let $q_z$ be a state, with
  $q_z=q_{z'}\Rightarrow z=z'$ for all $z,z'\in Z$. Then $T_\sigma=(Q,\allowbreak A,\allowbreak V_o,\allowbreak I,\allowbreak q_{w_0},\allowbreak \kn,\allowbreak \delta,\allowbreak \may,\allowbreak \safe,\allowbreak C)$, with
  $Q=\{q_z\mid z\in Z\}$, $\kn(q_z)=\{z\}$, $\may(q_z)=\sigma(z)$, and $\delta(q_z,a,z')=q_{z'}$ for every $z\in Z$,
  $a\in\sigma(z)$ and $z'\in\Post(z,a)$, undefined elsewhere, satisfies $T_\sigma\in\mathcal{T}$ and $\vdash T_\sigma$.
\item[(b)] Let $T=(Q,\allowbreak A,\allowbreak V_o,\allowbreak I,\allowbreak q_0,\allowbreak \kn,\allowbreak \delta,\allowbreak \may,\allowbreak \safe,\allowbreak C)\in\mathcal{T}$ with $\vdash T$, and
  \begin{align*}
  Z_T&=\{\,z\in W\mid \exists q\in\mathit{Reach}(T).\ \kn(q)=\{z\}\,\},\\
  \sigma_T(z)&=\bigcup_{q\in\mathit{Reach}(T),\ \kn(q)=\{z\}}\may(q).
  \end{align*}
  Then $(Z_T,\sigma_T)$ is a shield and $w_0\in Z_T$.
\end{itemize}
\end{theorem}
With full observation every belief is a single world. In (a), each shield world becomes a distinct typestate state
whose belief contains just that world, with the same permitted actions. In (b), the shield collects the worlds of
reachable states and, at each of them, the actions permitted at states with that belief. Where a typestate, depending
on its history, permits different actions at the same world in different states, the shield permits their union.

\begin{proposition}[Synthesis yields maximally permissive shield]\label{prop:shield}
Let $V_o=V$, $C$ be empty and $I=\{w_0\}$. Then $I\in X^\ast$ if and only if $w_0\in\mathit{Win}$, and every
$K\in X^\ast$ is $K=\{w\}$ for some $w\in\mathit{Win}$, with $\mathit{perm}(X^\ast,\{w\})=\mathit{shield}(w)$.
\end{proposition}
Synthesis thus succeeds when the initial world is winning, and at every surviving belief it permits
only the actions of the maximally permissive shield.

Epistemic typestates extend this core in two directions. With partial observation, they act on beliefs computed from
what the agent senses and remembers. Without declarations, synthesis then yields the winning region over the
reachable beliefs, as in shielding under partial observation~\cite{JungesJS21,CarrJJT23}. With declarations, agents
commit to actions that others can rely on through their public types (Section~\ref{sec:teams}).

\section{Evaluation}\label{sec:evaluation}

We evaluate epistemic typestates on nine benchmarks, measuring safety, task performance and the choices left to the learner. 

\paragraph{Benchmarks}
The nine benchmarks in Table~\ref{tab:benchmarks} cover a range of conditions. Most have
discrete worlds, but in MPE and VMAS positions and velocities are continuous, so the worlds are infinitely many and
synthesis cannot apply: there, as in three of the discrete benchmarks, types are built from guards, while the
remaining four are synthesised. Partial observation is the rule: in every benchmark except Connector and LBF, where agents see the whole board, each agent senses only part
of the world, and some of what it cannot sense matters for its safety. The benchmarks also differ in how agents
depend on each other. In Car Platoon the dependence runs along a chain, each follower depending on the cars ahead,
and in LBF it rests on declarations, as a player next to the food commits to loading it whenever another is too,
whereas in Traffic Junction, MPE and VMAS every agent can endanger every other. Teams range from two to ten agents,
and Car Platoon and LBF are run with two, five and ten agents to study scaling. Car Platoon appears in both of its
published versions, by Adalat et al.~\cite{AdalatHB26} and by Brorholt et al.~\cite{BrorholtLS25}, which differ in
speed range, gap bounds and episode length. Connector, Flatland, LBF, Pressure Plate\footnote{https://github.com/uoe-agents/pressureplate} and RWARE follow the
configurations of Adalat et al., and Traffic Junction uses the medium map of IC3Net~\cite{SinghJS19}.

\begin{table}[!t]
\centering
\caption{Benchmarks. Counts refer to controlled agents. Type inputs can be combined with own history and public
facts.}
\label{tab:benchmarks}
\footnotesize
\setlength{\tabcolsep}{3pt}
\renewcommand{\arraystretch}{1.05}
\begin{tabular}{@{}>{\raggedright\arraybackslash}p{.21\linewidth}
                  >{\raggedright\arraybackslash}p{.09\linewidth}
                  >{\raggedright\arraybackslash}p{.30\linewidth}
                  >{\raggedright\arraybackslash}p{.31\linewidth}@{}}
\toprule
\textbf{Benchmark} & \textbf{Agents} & \textbf{Safety requirement} & \textbf{Information used by the type} \\
\midrule
\multicolumn{4}{@{}l}{\emph{Discrete worlds, synthesised types}} \\
\addlinespace[2pt]
Car Platoon
& 2, 5, 10 & Every gap stays in $(0,20)$, or in $(0,200)$ in Brorholt et al.'s version
  & Own speed, predecessor speed and own gap \\
\addlinespace[2pt]
Connector~\cite{BonnetEtAl24}
& 2 & A reserved cell stays clear until its owner connects & Own head marker on the observed board \\
\addlinespace[2pt]
Flatland~\cite{MohantyEtAl20}
& 2 & No train is deadlocked & Own distance, stopped and opposite-train features \\
\addlinespace[2pt]
LBF~\cite{ChristianosSA20}
& 2, 5, 10 & No load attempt fails & Full-sight food and player observation \\
\midrule
\multicolumn{4}{@{}l}{\emph{Discrete worlds, types from guards}} \\
\addlinespace[2pt]
Pressure Plate
& 3 & Until the goal, the runner can return through every door it has reached & Own position \\
\addlinespace[2pt]
RWARE~\cite{ChristianosSA20}
& 2 & Unloaded robots in the queue yield to carriers of requested shelves
  & Own position, direction, load and highway bit \\
\addlinespace[2pt]
Traffic Junction~\cite{SinghJS19}
& 10 & No agent-caused crash & Native $3\times3$ window \\
\midrule
\multicolumn{4}{@{}l}{\emph{Continuous worlds, types from guards}} \\
\addlinespace[2pt]
MPE simple spread~\cite{LoweWTHAM17}
& 6 & No agent-caused collision & Own position and velocity; anonymous peer positions \\
\addlinespace[2pt]
VMAS navigation~\cite{BettiniKBP24}
& 6 & No collision & Own position, velocity and goal offset; 12 LIDAR rays \\
\bottomrule
\end{tabular}
\end{table}

\paragraph{Learners}
We use IPPO~\cite{DeWittEtAl20} and IQL~\cite{Tan93}, with the implementations supplied by Adalat et al., except in Brorholt et al.'s version of Car
Platoon, where we keep their UPPAAL cascading learner. Each learner is trained inside the typestate, and also
with the shield of Adalat et al.\ or Brorholt et al.\ in their versions of the benchmarks, where that shield can be
constructed, or without a shield in Traffic Junction. In MPE and VMAS, where neither shielding method applies because
both need a finite state space, the comparison is with Lagrangian versions of IPPO and IQL, trained without a mask: each agent receives as cost the onsets, i.e., first steps, of the agent-caused collisions it takes part
in, with cost limit~0. Barrier-certificate and reciprocal-velocity filters~\cite{WangAE17,BergGLM11} do not apply either: these agents choose among finitely
many actions and sense no velocity of the others. The typestate masks the
learner's actions and the learner keeps its original observations. Within a benchmark, the variants share the simulator, hyperparameters, rewards and training
budget, and, apart from the Lagrangian learners, differ only in the controller. All experiments ran on one machine,
an AMD Ryzen Threadripper PRO 5965WX, using 22 cores and 48\,GiB of memory. %

\paragraph{Training and evaluation}
Every IPPO and IQL run trains for 999936 environment steps (7812 updates of 128 steps), where in each step all
agents act once. IPPO and IQL are trained with five seeds, and their values are means $\pm$
standard errors over these seeds. Two means differ significantly when Welch's two-sided $t$-test at 5\% on the
per-run values rejects their equality. When both standard errors are 0, they differ significantly whenever they are unequal.
Given the five training runs per variant, statistical tests may have limited power to detect small differences. We
therefore report all means together with their standard errors.
Each trained policy is evaluated on 1000 episodes, always taking its highest-scoring permitted action.
Training seeds are fresh, and evaluation uses a separate seed block shared by all variants.

\paragraph{Assurance}
We check the types used in the experiments independently of the engine that builds them. Every type of the discrete
benchmarks is exported with its model and verified by two checkers that share no code with the engine, one of them
proved sound in Rocq: a typestate passes only if it is well formed, and a guard only if it
satisfies the hypothesis of Proposition~\ref{prop:guard}. All 34 exported types pass. Except in Traffic Junction,
the exported typestates also carry a certificate, verified by both checkers, that at every reachable state they
permit exactly the actions of the synthesised typestate. They are therefore maximally
permissive in the sense of Proposition~\ref{prop:greatest}. To test the checkers themselves, we build 194 corrupted
copies of the exports, e.g., by permitting an excluded action, dropping a transition or adding an unsafe
successor, and both checkers reject every one.

Theorem~\ref{thm:safety} also requires each agent's model to cover what the simulator does while the others act
within their public types. We test this hypothesis exhaustively on the unchanged simulators of Car Platoon with two
followers, Connector, Flatland, LBF with two agents and Pressure Plate. From the public reset, which is the same for
every seed of the experiments, the check visits every reachable team state and executes there every joint action the
types permit, under every outcome of the simulator's random draws: in 104 (Flatland) to 81267 (Connector) team
states, the types predict every observation and no step is unsafe. Theorem~\ref{thm:safety} involves no check of the whole team: its hypotheses concern each
agent's type, its model and the public types it relies on, and the acyclicity of reliance on safety after the step.
Adding agents therefore adds no obligation as long as it changes none of these. It changes none in Car Platoon and
LBF, whose models see the others only through the cars ahead and through whether two players are next to the food,
and whose five- and ten-agent teams run the very types of the checked two-agent instances. In every run of these
teams, each typed controller also checks at every step that its agent's observation is one its type predicts, and
stops the run otherwise (no run stopped). In Traffic Junction, a similar check is exhaustive on the easy map with three to five
cars, and on the medium map the model agrees with the simulator on all 40000 transitions of 1000 sampled episodes.
For MPE and VMAS, the hypothesis of Proposition~\ref{prop:guard} rests on a written argument, an invariant on pairs of
agents that permitted actions preserve. Its numerical inputs include the dynamics model, checked against the simulator
on sampled steps, and, in VMAS, a braking-distance bound by interval arithmetic and a reduction to finitely many ray
directions. These types are thus argued and tested rather than certified, so we try to refute them. Every agent runs
its type on its own observations, while an
adversarial search chooses the initial positions and, among each agent's permitted actions, the one that most
reduces its distance to another agent, using the true state. In 8000 MPE and 7845 VMAS episodes, with 1.2 and 4.7
million decisions, the search finds no agent-caused collision. Its closest approaches, 0.3005 and 0.2143 against
collision distances of 0.3 and 0.205, show that it brings agents close to colliding.

\subsection{Discrete benchmarks}\label{sec:discrete}

No typed learner causes a safety violation, in training or in evaluation, in any benchmark, seed or learner, and
neither does any shielded learner. Unshielded learners in Traffic Junction cause $100554\pm2133$
(IPPO) and $182115\pm1251$ (IQL) agent-caused crash onsets per training run. We report undiscounted team return and success where the benchmark defines it.

\paragraph{Adalat et al.'s shields}
Neither typed learners nor learners with Adalat et al.'s shield~\cite{AdalatHB26} achieve a significantly higher team
return (Table~\ref{tab:discrete-adalat}), except in Car Platoon with IPPO, where the shield reads state that no agent
can sense. Returns are equal on every seed in five of the twelve comparisons, and the other differences go both ways.
\begin{table}[!htb]
\centering
\caption{Performance on Adalat et al.'s benchmarks ($n$: agents). Adalat mask local: whether own history and contract determine the shield's mask, in Flatland only for the profiles selected for evaluation. $^{*}$Checked on winning states. \textsc{nc}: no certified profile found. \textsc{budget}: not completed within 32\,GiB. Car Platoon reports negative gap rewards.}
\label{tab:discrete-adalat}
\footnotesize
\setlength{\tabcolsep}{4pt}
\renewcommand{\arraystretch}{1.08}
\begin{tabular}{@{}lcrrrr@{}}
\toprule
\textbf{Benchmark ($n$)} & \shortstack{\textbf{Adalat}\\\textbf{mask}\\\textbf{local?}} & \multicolumn{2}{c}{\textbf{IPPO}} & \multicolumn{2}{c}{\textbf{IQL}} \\
\cmidrule(lr){3-4}\cmidrule(l){5-6}
 & & \textbf{Typestate} & \textbf{Adalat} & \textbf{Typestate} & \textbf{Adalat} \\
 & & \multicolumn{4}{c}{Team return $\uparrow$} \\
\midrule
Car Platoon, 2 & No & $-464.6\pm0.0$ & $-419.8\pm0.0$ & $-464.6\pm0.0$ & $-449.4\pm8.2$ \\
Car Platoon, 5 & -- & $-1528.9\pm0.0$ & \textsc{nc} & $-1533.8\pm4.9$ & \textsc{nc} \\
Car Platoon, 10 & -- & $-4206.3\pm0.0$ & \textsc{nc} & $-4206.4\pm0.1$ & \textsc{nc} \\
Connector, 2 & Yes & $1.7\pm0.0$ & $1.7\pm0.0$ & $1.2\pm0.3$ & $1.2\pm0.3$ \\
Flatland, 2 & Profile & $10.8\pm0.0$ & $10.8\pm0.0$ & $10.8\pm0.0$ & $8.6\pm1.7$ \\
LBF, 2 & Yes$^{*}$ & $1.0\pm0.0$ & $1.0\pm0.0$ & $0.0\pm0.0$ & $0.0\pm0.0$ \\
LBF, 5 & -- & $1.0\pm0.0$ & \textsc{budget} & $0.8\pm0.2$ & \textsc{budget} \\
LBF, 10 & -- & $1.0\pm0.0$ & \textsc{budget} & $1.0\pm0.0$ & \textsc{budget} \\
Pressure Plate, 3 & No & $14.2\pm0.1$ & $14.3\pm0.0$ & $14.3\pm0.0$ & $14.3\pm0.0$ \\
RWARE, 2 & No & $0.80\pm0.20$ & $0.20\pm0.20$ & $0.60\pm0.24$ & $1.00\pm0.00$ \\
\bottomrule
\end{tabular}
\end{table}

\begin{figure}[!htb]
\centering
\includegraphics[width=0.9\linewidth]{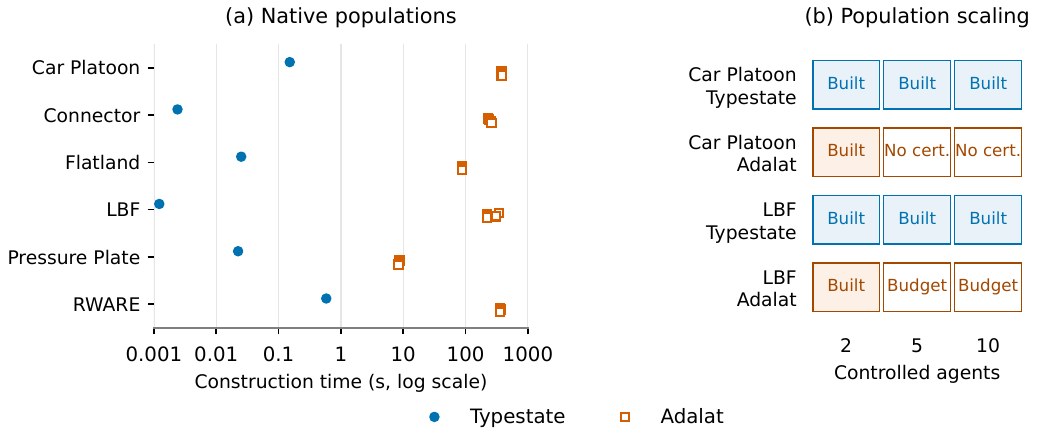}
\caption{Construction time and availability. (a)~Times at the native populations: each Adalat mark is one of five attempts, including the profile search, and each typestate mark one
local compilation. (b)~Outcomes at two, five and ten agents. ``No cert.'': no certified profile found. ``Budget'':
not completed within 32\,GiB.}
\label{fig:adalat-construction}
\end{figure}

For the evaluated shields in
Car Platoon, Pressure Plate and RWARE, we prove that no agent can reconstruct its exact mask from its complete
local history,
even with the active contract known: two reachable executions indistinguishable to that agent receive
different masks, so there the shield cannot run on the agents' own information.\footnote{Histories include observations, own actions and rewards, states of the agent's
contract monitor, and all earlier masks.} In Flatland, one profile used during training has this nonlocal
dependence, but all profiles selected for final evaluation are locally recoverable. In Car Platoon with
two followers, where the team return is the negative accumulated gap cost, the typestate incurs 10.7\% more cost than
Adalat et al.'s shield with IPPO, consistent with the significant difference in team return reported above. With
IQL, the cost is 3.4\% higher, but the difference is not statistically significant. In RWARE, where every run's return is 0 or 1, typed learners earn 1
in four (IPPO) and three (IQL) of five runs, against one and five with the shield. In Pressure Plate,
our types also match the shield's $100\%$ success using only own position, the public layout and public plate ownership.
Construction also determines which team sizes can be evaluated (Figure~\ref{fig:adalat-construction}).
Adalat et al.'s construction searches contract profiles, exponentially many in the number of agents, for ones it
can certify over the global model, whereas ours compiles each agent's local model against the public types of
the others. At the native populations, local compilation takes $0.001$--$0.58$\,s, against $8.4$--$379.5$\,s for
theirs. The same two Car Platoon roles and single LBF role serve teams of two, five and ten agents. At five and
ten followers, none of the 12000 profiles examined with Car Platoon settings is certified, in any of
five attempts. LBF construction at five and ten agents does not complete within 32\,GiB.

\paragraph{Brorholt et al.'s shields}
Neither controller has a significantly lower summed follower cost at two, five or ten followers
(Table~\ref{tab:discrete-brorholt}): the type's cost differs from the shield's by $-1.0\%$, $-2.0\%$ and $+2.1\%$. Both controllers read
only own speed, predecessor speed and own gap, and we change only the action mask supplied to the authors'
UPPAAL cascading learner~\cite{BrorholtLS25}. Each of ten repetitions per fleet size uses 6667 training
episodes per follower and 1000 evaluation episodes. An independent exact check on the authors' state
lattice confirms safety for both controllers. 
The recorded local type build takes $4.5$\,s, against $14.1$\,s on average for the shield. Both methods reuse
one local controller at every follower position.

\begin{table}[!htbp]
\centering
\caption{Car Platoon with Brorholt et al.'s cascading learner and the same local observations, as means $\pm$ standard errors over ten repetitions.}
\label{tab:discrete-brorholt}
\footnotesize
\setlength{\tabcolsep}{8pt}
\renewcommand{\arraystretch}{1.08}
\begin{tabular}{@{}rrr@{}}
\toprule
\textbf{Followers} & \textbf{Typestate} & \textbf{Brorholt} \\
 & \multicolumn{2}{c}{Summed follower cost $\downarrow$} \\
\midrule
2 & $5526.6\pm78.3$ & $5580.5\pm82.3$ \\
5 & $14669.0\pm215.3$ & $14961.8\pm221.6$ \\
10 & $32146.7\pm747.6$ & $31477.7\pm383.6$ \\
\bottomrule
\end{tabular}
\end{table}

\paragraph{Traffic Junction}
Cars use their native $3\times3$ window, own history and public road layout. After braking, the type requires
GAS whenever GAS is permitted. Table~\ref{tab:discrete-traffic} separates agent-caused crashes from spawn collisions.
Typed learners record no agent-caused crash onsets, against $2.24\pm0.06$ per episode for unshielded IPPO and
$0.33\pm0.01$ for unshielded IQL. Spawn collisions remain possible, and are more frequent with the type: the simulator can place a new car on an occupied
entry cell. IPPO's success and raw return improve. With IQL, native success is lower with the type, but raw return is
higher and completed routes increase to $11.59\pm0.39$ per episode. Native success counts spawn collisions
as failures. Excluding only spawn penalties gives higher returns: for IQL, $-26.6\pm0.7$ against
$-56.1\pm0.7$. Brorholt et al.'s ordered composition does not cover the
departure rules that these cars rely on in each other's public types.

\begin{table}[!htbp]
\centering
\caption{Traffic Junction on the medium map, as means $\pm$ standard errors over five training seeds. Success is the native fraction of collision-free episodes. Spawn collisions also count as failures. Crash columns count onsets per episode.}
\label{tab:discrete-traffic}
\footnotesize
\setlength{\tabcolsep}{4pt}
\renewcommand{\arraystretch}{1.08}
\begin{tabular}{@{}lrrrrr@{}}
\toprule
\textbf{Learner/type} & \textbf{Success} & \textbf{Return} & \shortstack{\textbf{Return without}\\\textbf{spawn penalties}} & \shortstack{\textbf{Agent}\\\textbf{crashes}} & \shortstack{\textbf{Spawn}\\\textbf{crashes}} \\
\midrule
\shortstack[l]{IPPO\\Typestate} & $0.65\pm0.00$ & $-32.9\pm0.0$ & $-19.5\pm0.0$ & $0.00\pm0.00$ & $0.57\pm0.00$ \\
\shortstack[l]{IPPO\\Unshielded} & $0.11\pm0.01$ & $-121.5\pm2.1$ & $-121.5\pm2.1$ & $2.24\pm0.06$ & $0.00\pm0.00$ \\
\addlinespace
\shortstack[l]{IQL\\Typestate} & $0.52\pm0.01$ & $-51.0\pm0.8$ & $-26.6\pm0.7$ & $0.00\pm0.00$ & $0.84\pm0.01$ \\
\shortstack[l]{IQL\\Unshielded} & $0.72\pm0.01$ & $-56.9\pm0.9$ & $-56.1\pm0.7$ & $0.33\pm0.01$ & $0.03\pm0.01$ \\
\bottomrule
\end{tabular}
\end{table}

\subsection{Continuous benchmarks}\label{sec:continuous}

No typed learner causes a safety violation, while Lagrangian ones cause $17231\pm67$ (IPPO) and $35382\pm597$ (IQL)
agent-caused collision onsets per training run in MPE, and $2540\pm16$ and $2735\pm18$ in VMAS
(Table~\ref{tab:continuous}). We also report the targets reached at the end of an episode: the
landmarks covered in MPE and the robots on their goal in VMAS, out of six.

\begin{table}[!htbp]
\centering
\caption{Continuous benchmarks, as means $\pm$ standard errors over five training seeds. Targets are the landmarks covered (MPE) and the robots on their goal (VMAS) at the end of an episode, out of six. Collisions are agent-caused onsets.}
\label{tab:continuous}
\footnotesize
\setlength{\tabcolsep}{4pt}
\renewcommand{\arraystretch}{1.08}
\begin{tabular}{@{}llrrrr@{}}
\toprule
\textbf{Learner} & \textbf{Variant} & \textbf{Return} $\uparrow$ & \textbf{Targets} $\uparrow$ & \shortstack{\textbf{Collisions}\\\textbf{per episode} $\downarrow$} & \shortstack{\textbf{Collisions}\\\textbf{in training} $\downarrow$} \\
\midrule
\multicolumn{6}{@{}l}{\emph{MPE simple spread}} \\
\addlinespace
IPPO & Typestate & $-180.1\pm5.4$ & $0.318\pm0.029$ & $0.000\pm0.000$ & $0\pm0$ \\
 & Lagrangian & $-355.4\pm2.3$ & $0.026\pm0.002$ & $0.035\pm0.002$ & $17231\pm67$ \\
\addlinespace
IQL & Typestate & $-233.1\pm10.4$ & $0.178\pm0.023$ & $0.000\pm0.000$ & $0\pm0$ \\
 & Lagrangian & $-268.0\pm9.1$ & $0.064\pm0.006$ & $0.343\pm0.070$ & $35382\pm597$ \\
\midrule
\multicolumn{6}{@{}l}{\emph{VMAS navigation}} \\
\addlinespace
IPPO & Typestate & $30.2\pm0.2$ & $2.561\pm0.189$ & $0.000\pm0.000$ & $0\pm0$ \\
 & Lagrangian & $31.6\pm0.3$ & $1.136\pm0.071$ & $0.000\pm0.000$ & $2540\pm16$ \\
\addlinespace
IQL & Typestate & $17.7\pm0.9$ & $0.235\pm0.031$ & $0.000\pm0.000$ & $0\pm0$ \\
 & Lagrangian & $-11.2\pm6.2$ & $0.062\pm0.018$ & $0.022\pm0.012$ & $2735\pm18$ \\
\bottomrule
\end{tabular}
\end{table}

\paragraph{MPE simple spread}
Each agent observes its own position and velocity and the anonymous positions of the five others, but never their
velocities. From its own history it recovers their previous velocities, and their current ones up to their last
action. Agents can overlap without any of them causing it: the reset
places $0.98$ overlapping pairs per episode on average, and contact forces push agents into further overlaps. With
the type, these occur only in episodes that start with an overlapping pair, $0.23\pm0.01$ (IPPO) and $0.34\pm0.03$
(IQL) per episode. The Lagrangian learners incur $0.12\pm0.00$ and $0.23\pm0.02$. A collision is agent-caused when, at some step
since the last contact of either agent or since the reset, both agents taking no-op from then on would have kept them
apart (the type always permits no-op). It permits a deviation only if
each pair the agent belongs to remains, at every remaining step, at least as far apart as if both took no-op now, or
at least a threshold apart: in each pair, one agent checks this against its partner's no-op and the other against
every action the type permits the partner. Typed IPPO and IQL obtain higher returns and cover more landmarks than
their Lagrangian versions. After training, the Lagrangian learners still cause collisions in $3.4\%$ (IPPO) and
$28.1\%$ (IQL) of evaluation episodes.

\paragraph{VMAS navigation}
Each robot observes its own position, velocity and goal offset, and twelve LIDAR rays that detect other robots up to
a range of $0.35$. Robots start at least $0.25$ apart, so every collision is agent-caused. A robot's belief places
every other robot outside the region its rays show free. Besides always permitting braking, the typestate permits another action
only if, as the robot brakes afterwards, it advances towards each possible position of another robot by at most half
of the room beyond a fixed clearance, and either stays farther than the clearance plus the distance that robot can
travel while braking, or moves away from it. The public type of every robot permits only thrusts that keep each
velocity component within $0.25$, which bounds that distance. With IQL, typed learners obtain higher returns and bring more robots to their
goal. With IPPO, the Lagrangian learner's return is higher, $31.6\pm0.3$ against $30.2\pm0.2$, while typed learners
bring more robots to their goal, $2.56\pm0.19$ against $1.14\pm0.07$. The return rewards progress
towards the goal at every step. In evaluation, Lagrangian IPPO causes no collision, and Lagrangian IQL causes
collisions in $2.1\%$ of episodes.

\subsection{Freedom left to the learner}\label{sec:freedom}

\begin{figure}[!htbp]
\centering
\includegraphics[width=0.75\linewidth]{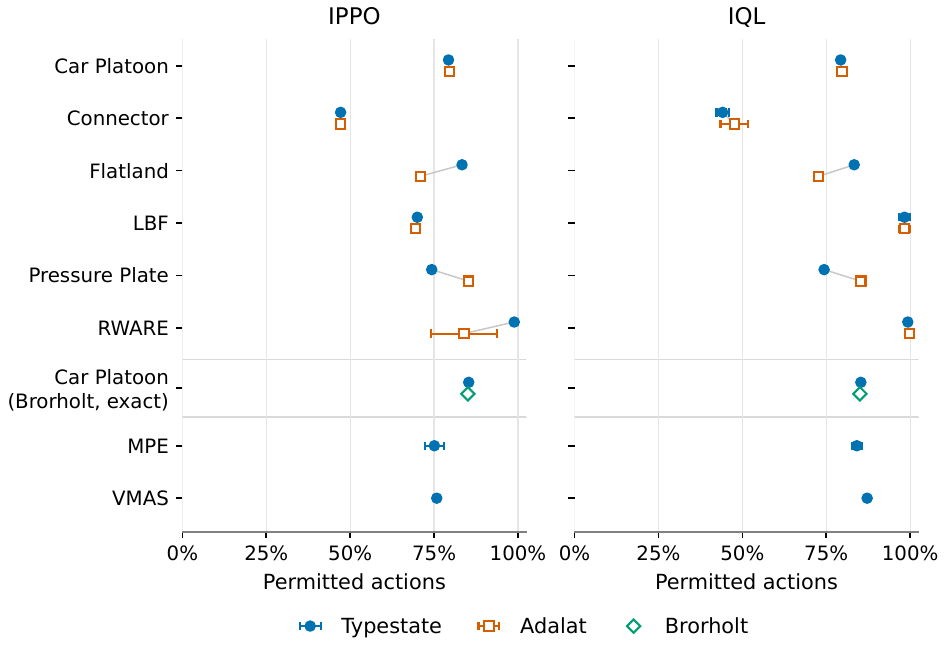}
\caption{Shares of actions retained by types and shields. Except for the Brorholt row, points summarise evaluation
decisions and bars show one standard error over five seeds. The Brorholt row averages uniformly over reachable states under both controllers. }
\label{fig:freedom}
\end{figure}

Freedom is the share of actions left to the learner. 
Figure~\ref{fig:freedom} compares the freedom retained by types and both shielding methods in the discrete
benchmarks, and by types in the continuous ones. Against Adalat et al., typed learners can retain more actions or
fewer. In Flatland, they retain $83.3\pm0.0\%$ with both learners, against $70.9\pm0.5\%$ (IPPO) and $72.7\pm0.3\%$
(IQL) with the shield. In Pressure Plate, they retain $74.3\pm0.1\%$ (IPPO) and $74.4\pm0.0\%$ (IQL), against $85.3\pm0.0\%$.
In Traffic Junction, typed IPPO retains $88.8\pm0.0\%$ of actions for cars on the road, and IQL retains
$76.3\pm1.0\%$. 
For Brorholt et al., the plotted shares average equally over the same reachable states: the type
retains $85.31\%$ of actions, against $85.05\%$ with the shield. This comparison also shows that the
type retains every shield-permitted action and allows more in 91 states. 

In the continuous benchmarks, the Lagrangian learners act without a mask (thus, they have no entry in Figure~\ref{fig:freedom}). Typed learners retain $75.1\pm2.9\%$
(IPPO) and $84.1\pm1.4\%$ (IQL) of actions in MPE, and $75.8\pm0.6\%$ and $87.2\pm0.4\%$ in VMAS. The types remove
at least one action in $59.2\pm5.5\%$ (IPPO) and $40.2\pm3.4\%$ (IQL) of decisions in MPE, and in $51.0\pm0.8\%$ and
$31.8\pm0.8\%$ in VMAS. In the other decisions, every action is permitted. In VMAS, the removed actions include
every thrust that would take a velocity component beyond $0.25$. Freedom is measured on each
learner's decisions, so it depends on the states visited: in both continuous benchmarks, IPPO
retains fewer actions than IQL, whereas in Traffic Junction IQL retains fewer.

\section{Discussion and limitations}\label{sec:discussion}

Epistemic typestates give learners the guarantee of a shield while lifting restrictions of existing shielding methods.
Against Adalat et al., they scale: the same roles serve teams of two, five and ten agents, whereas at five and ten agents, Adalat et al.'s construction finds
no certified profile or does not complete within 32\,GiB. Against Brorholt et al., they are more expressive: theirs compose
memoryless shields in a fixed order, whereas typestates let agents rely on each other's public types in both directions 
and use the agent's own history. Against both, types
need no finite state space: both shielding methods assume one, and Brorholt et al.\ leave continuous systems to future
work, while guards yield types for the infinitely many worlds of MPE and VMAS. Task
performance does not differ significantly from that of either shielding method in all comparisons but one, against
Adalat et al.'s shield in Car Platoon, where it reads state that no agent can sense.

Freedom explains why the shields do no better. No well-formed typestate permits more than a synthesised one given the same
belief, except where an obligation fixes the action (Proposition~\ref{prop:greatest}), and the certificates of
Section~\ref{sec:evaluation} show that every exported typestate outside Traffic Junction permits exactly what synthesis
would. Outside obligations, a shield can therefore permit more than such a type only by knowing or assuming more
than the type, as Adalat et al.'s shield does in Car Platoon. With the same information, Brorholt et al.'s shield
permits nothing that the type withholds, and the type permits more in 91 states. In Traffic Junction,
whose guard is not certified maximal, the type comes close: on the easy map, with three to five cars, it agrees with an
exact reference that recalls the car's whole observation history in all but 8 decisions, where it withholds
GAS that the reference permits.

Because well-formedness is checked state by state (Definition~\ref{def:wf}) and each agent separately
(Theorem~\ref{thm:safety}), an independent checker proved sound in Rocq can certify the finite types before they are
used.

Finally, a type constrains the learner only through its action mask. Each type serves IPPO and IQL alike, as well
as Brorholt et al.'s cascading learner in their version of Car Platoon, and none of these learners has to cause a
violation to learn to avoid it. Constrained learners, by contrast, learn from the cost of the violations they cause.

\subsection{Limitations}\label{sec:limitations}

Like every model-based safety method, epistemic typestates guarantee safety only as far as their models are right:
Theorem~\ref{thm:sound} assumes that the environment follows the specified world dynamics, and
Theorem~\ref{thm:safety} that each model covers what the others may do within their public types. Writing these
models remains the designer's task, although the effort grows with the kinds of agents rather than with their number:
each benchmark declares one or two roles, which serve all its agents. The team checks of Section~\ref{sec:evaluation} test them against the simulators, but testing cannot prove a model sound, and the certification of exported types trusts
the exporter that produces them.

Synthesis requires finitely many reachable beliefs, and its cost grows with their number, which in the worst case is
exponential in the number of worlds. The cost depends on each agent's local model rather than on the joint behaviour
of the team, but it limits how much of the world a synthesised type can afford to track. When the worlds are
infinite, types come from guards, which do not enumerate
beliefs but carry no maximality guarantee of their own, and whose hypothesis must be established for each model. For
continuous dynamics, we currently establish it by numerically checked arguments rather than with a verified checker. The formalism also keeps a finite action set, so continuous actions remain outside its current scope.

Outside obligations, a synthesised type is maximally permissive for the information it uses, and choosing that information is a design
decision: tracking more observations or more history can permit more, but may enlarge the set of beliefs that
synthesis must explore. The engine's analyses, such as the comparison with an exact reference in
Section~\ref{sec:discussion}, measure what a coarser choice costs. Finally, agents may rely on each other's public types
in both directions, but reliance on safety after the step must be acyclic (Theorem~\ref{thm:safety}), and public types
must be agreed before deployment: the approach replaces communication at run time with agreement at design time.

\section{Conclusion}\label{sec:conclusion}

We have introduced epistemic typestates, a type discipline for safe multi-agent RL. Agents learn inside types that
permit only the actions whose safety follows from what each agent knows, so safety holds for every learner, in
training as in execution, without communication. Well-formed types keep their agents safe and compose into team
safety. Synthesis finds a well-formed type whenever one exists, as permissive as any other apart from declared
obligations, and guards give types for infinitely many worlds. Shields are the special case with full observation and
no declarations, and all proofs are mechanised in Rocq. On all benchmarks, no typed learner causes a safety violation,
and task performance does not differ significantly from that of two recent multi-agent shielding methods in all
comparisons but one, although one of them reads state that no agent can sense in three benchmarks. In Adalat et al.'s benchmarks, types for ten agents are built in under a second, whereas their construction
found no shield for teams of five or ten agents. Types also express mutual commitments and history-dependent rules,
which ordered memoryless shields cannot, and extend to continuous worlds, where Lagrangian learners cause thousands of
violations in every training run.

Several directions remain open. A refinement relation relative to sensing would let a type be replaced by one that permits less
and commits to more, as far as the other agents can tell, without checking the team again. Constructing and verifying
guards automatically would bring continuous dynamics within reach of the certified checker, and continuous action sets
would extend the formalism beyond finite choices. Learning models from data would reduce the main burden that the
approach places on the designer.

\section*{Availability}

All code, proofs and results, including raw outputs, are archived
on Zenodo under the MIT licence: \url{https://doi.org/10.5281/zenodo.24371708}.

\bibliographystyle{elsarticle-num}
\bibliography{references}

\end{document}